\documentclass[11pt,titlepage]{article}
\usepackage[margin=1.2in]{geometry}
\usepackage{amsmath, amsbsy, amsthm, array, mathrsfs}
\usepackage{graphics}
\usepackage{physics}
\usepackage{epsfig, amssymb,latexsym,verbatim}
\usepackage{graphicx}
\usepackage{color}
\usepackage{dsfont, bbm}
\usepackage{relsize}
\usepackage{subcaption}

\newcommand{\R}{\mathbb{R}}

\newcommand{\noin}{\noindent}
\newcommand{\bee}{\begin{eqnarray*}}
\newcommand{\ene}{\end{eqnarray*}}
\newcommand{\bec}{\begin{center}}
\newcommand{\enc}{\end{center}}
\newcommand{\be}{\begin{equation}}
\newcommand{\ee}{\end{equation}}

\newcommand{\mc}{\mathcal}

\newcommand{\ep}{\varepsilon}
\newcommand{\mb}{\mathbf}
\newcommand{\bs}{\boldsymbol}
\newcommand{\tb}{\textbf}
\newcommand{\pend}{$\blacksquare$}
\newcommand{\vs}{\vskip 3mm}
\newcommand{\bi}{\begin{itemize}}
\newcommand{\ei}{\end{itemize}}

\begin{document}

\title{\LARGE  Robust penalized least squares of depth trimmed residuals regression  
for high-dimensional data}
\vs
\vs
\author{ {\sc 
Yijun Zuo}\\[2ex]
         {\small {
         \em Department of Statistics and Probability} }\\[.5ex]
         {\small Michigan State University, East Lansing, MI 48824, USA} \\[2ex]
         {\small 
         zuo@msu.edu}\\[6ex]
     }
 \date{\today}
\maketitle

\vskip 3mm
{\small

\begin{abstract}
Challenges with data in the big-data era include (i) the dimension $p$ is often  larger than the sample size $n$ (ii) outliers or contaminated  points are frequently hidden and more difficult to detect. Challenge (i) renders most conventional methods inapplicable. Thus, it attracts tremendous attention from statistics, computer science, and bio-medical communities.  Numerous penalized regression methods have been introduced as modern methods for analyzing high-dimensional data. Disproportionate attention has been paid to the challenge (ii) though.
Penalized regression methods can do their job very well and are expected to handle the challenge (ii) simultaneously. The fact is most of them can break down by a single outlier (or single adversary contaminated point) as revealed in this article.
 The latter systematically examines leading penalized regression methods in the literature in terms of their robustness and provides  quantitative assessment and reveals that most of them can break down by a single outlier. Consequently, a novel robust penalized regression method based on the least sum of squares of depth trimmed residuals is proposed and studied carefully.
Experiments with simulated and real data reveal that the newly proposed method can outperform some leading competitors in terms of estimation and prediction accuracy in the cases considered.
\vs

\bigskip
\par
\vspace{9pt}

\noindent{\bf Key words and phrase:} penalized regression,  robustness of regularized regression estimators, least squares of depth trimmed residuals,  finite sample prediction error bound, approximate computation algorithms.
\bigskip
\par
\noindent {\bf Running title:} robust penalized  depth trimmed regression
\end{abstract}
}
\setcounter{page}{1}

\section {Introduction}

\noin
\tb{Least squares regression, the classical setting}~~
Consider the classic linear regression model,
\be y_i=(1, \bs{x}'_i)\bs{\beta}_0+e_i:=\bs{w}'_i\bs{\beta}_0+e_i, \label{model-population.eqn}
\ee
where random variables $y_i$ and $e_i$ $\in \R$, random vector $\bs{x}_i\in \R^{p-1}$, and $\bs{\beta}_0 \in \R^p$ is an unknown parameter of interest, $'$ stands for the transpose. One wants to estimate the $\bs{\beta}_0$ based on a given sample $\mb{Z}^{(n)} :=\{(\bs{x}'_i, y_i)', i\in\{1,\cdots, n\}\}$ from a parent model $y=\bs{w}'\bs{\beta}_0+e$. 
\vs
Call the difference between $y_i$ (observed value) and $\bs{w'_i}{\bs{\beta}}$ (predicted value),
 $r_i$, for a candidate coefficient vector $\bs{\beta}$ (which is often suppressed). 
\be r_i:= {r}_i(\bs{\beta})= y_i-(1,\bs{x}'_i)\bs{\beta}:=y_i-\bs{w'_i}{\bs{\beta}}.\label{residual.eqn}
\ee
To estimate $\bs{\beta}_0$, the classic \emph{least squares} (LS) estimator is the minimizer of the sum of the squared residuals (SSR):
$\widehat{\bs{\beta}}_{ls}=\arg\min_{\bs{\beta}\in\R^p} \sum_{i=1}^n r^2_i. $
Alternatively, one can replace the square by absolute value to obtain the least absolute deviations (lad)  estimator (aka, $L_1$ estimator, in contrast to the $L_2$ (LS) estimator).
A straightforward calculus derivation leads to
\be\widehat{\bs{\beta}}_{ls}=(\bs{X}'_n \bs{X}_n)^{-1}\bs{X}'_n\bs{Y}_n. \label{ls.eqn}
\ee
where $\bs{Y}_n=(y_1,\cdots, y_n)'$, $\bs{X}_n=(\bs{w}_1,\cdots, \bs{w}_n)'$ and the columns of $\bs{X}_n$ are assumed to be linearly independent (i.e. $\bs{X}_n$ has a full rank $p$ ($n\geq p$)).
\vs
The LS estimator is popular in practice across a broader spectrum of disciplines due to  its (i) great computability (with the computation formula);
   and (ii) optimal properties (the best linear unbiased estimator (BLUE) and the uniformly minimum variance  unbiased estimator (UMVUE), page 186 of \cite{S03} when the i.i.d. error $e_i$ follows a normal $\mc{N}(0,\sigma^2)$. 
   \vs
It, however, can behave badly when the error distribution is slightly departed from the normal distribution,
particularly when the errors are heavy-tailed or contain outliers.

\vs
\noin
\tb{Penalized regression, the state of the art}~~
In modern applied data analysis,  the number of variables often is even larger than the number of observations. Traditional methods such as LS can then no longer be applied due to
the design matrix $\bs{X}$ being less than $p$ rank ($n< p$), hence LS estimator is no longer unique and its variance 
is large if $\bs{X}$ is close to collinear.  Furthermore, 
models that include the full set of explanatory variables often have poor prediction
performance as they tend to have large variance while large models
are in general difficult to interpret.
\vs
\emph{Ridge regression},  minimizing SSR, subject to a constraint $\sum_{i=1}^p|\beta_i|^2<t$ \vspace*{-3mm}
\be \widehat{\bs{\beta}}_{ridge} (\lambda):= \arg\min_{\bs{\beta}\in \R^p}\Big\{\sum_{i=1}^n r^2_i+ \lambda\sum_{i=1}^p|\beta_i |^2\Big\},  \label{ridge.eqn}
\ee
 first proposed by \cite{HK70a, HK70b}, 
 is a useful tool for improving prediction in regression
situations with highly correlated predictors  and tackling the non-inverse issue,
\be\widehat{\bs{\beta}}_{ridge} (\lambda)=(\bs{X_n}'\bs{X_n}+\lambda I_{d \times d})^{-1}\bs{X_n}'\bs{Y_n}, \label{ridge-estimator.eqn}\ee
its variance is smaller than that of the LS estimator. Therefore, better estimation can be
achieved on the average in terms of mean squared error (MSE) with a little sacrifice of bias, and predictions can be improved overall.
\vs
The ridge regression was
generalized in \cite{FF93} that  introduced \emph{bridge regression},
which minimizes SSR subject to a constraint $ \sum_{i=1}^p|\beta_i|^{\gamma} \leq t$ with $\gamma \geq 0$, \vspace*{-2mm}
\be
\widehat{\bs{\beta}}_{bridge} (\lambda, \gamma):=\arg\min_{\bs{\beta}\in \R^p}\Big\{\sum_{i=1}^n r^2_i+ \lambda\sum_{i=1}^p|\beta_j |^{\gamma}\Big\}. \label{bridge.eqn}
\ee 
Ridge
regression ($\gamma = 2$) and subset selection  ($\gamma = 0$) are special cases.
\vs
 \emph{Least absolute
shrinkage and selection operator} (lasso) was introduced in \cite{T96},  minimizing SSR subject
to a constraint $ \sum_{j=1}^p|\beta_j| \leq t$, is a special case of the bridge with $\gamma = 1$. As pointed out
by \cite{T96}, the lasso shrinks the LS estimator $\widehat{\bs{\beta}}_{ls}$ towards $0$ and potentially sets
$\widehat{\beta}_j=0$ for some $j$. That is, it performs as a variable selection operator.
\vs
Other approaches of regularized regression include, among others, \tb{(i)} \cite{SO11}, 
 who
proposed an iterative procedure for outlier detection and consider the model
$y_i = \sum_{j=1}^p x_{ij}\beta_j +
\gamma_i + \epsilon_i,
$
in which the parameter $\gamma_i$ is nonzero when observation i is an
outlier. An earlier mean-shift model was proposed by \cite{MW10}. 
 \tb{(ii)} \emph{elastic nets}, 
  introduced in \cite{ZH05}, a generalization of the ridge and lasso models, which combines the two penalties and yields 
\be
\widehat{\bs{\beta}}_{enet}(\lambda_1, \lambda_2):=\arg\min_{\bs{\beta}\in \R^p}\Big\{\sum_{i=1}^n r^2_i  +\lambda_1\sum_{i=1}^p|\beta_i|+\lambda_2\sum_{i=1}^p\beta^2_i\Big\}. \label{elastic-nets.eqn}
\ee
\tb{(iii)}
To avoid the pre-estimation of standard deviation $\sigma$ of the error term in lasso and achieve a better performance, 
\emph{square-root lasso}, introduced in \cite{BCW11}, is defined as
\be
\widehat{\bs{\beta}}_{sqrt-lasso}=\arg\min_{\bs{\beta}\in \R^p}\Big\{ \big(\sum_{i=1}^n r^2_i\big)^{1/2}+\lambda \sum_{i=1}^p|\beta_i| \Big\}. \label{squre-rrot-lasso.eqn}
\ee
\tb{(iv)} Aim to control the false discover rate (FDR), 
  \emph{slope} (Sorted L-One Penalized Estimation) introduced in \cite{BVSSC15},
\be
\widehat{\bs{\beta}}_{slope}=\arg\min_{\bs{\beta}\in \R^p}\Big\{\sum_{i=1}^n r^2_i + \sum_{i=1}^p\lambda_i|\beta_{(i)}|\Big\},
\ee
where $\lambda_1\geq\lambda_2\geq \cdots\geq\lambda_p\geq 0$ and $|\beta_{(1)}|\geq |\beta_{(2)}|\geq \cdots\geq |\beta_{(p)}|$.
\vs
 Strong connections between some
modern methods and a method called \emph{least angle regression} (lar) was revealed in \cite{EHJT04} where they developed an
algorithmic framework that includes all of these methods (lasso, boosting, forward stagewise regression) and provided a fast implementation, for which they used the term ‘lars’.
lars is a promising technique/algorithm for variable selection applications, offering a nice alternative to stepwise regression.
  For an excellent review on lars and lasso, see 
  \cite{HCMF08}. 
\vs
Other outstanding penalized regression estimators include, among others, SCAD  \cite{F97}, 
\cite{FL01} 
and MCP \cite{Z10}. 
It is not our goal to review all existing penalized/regularized regression estimators in the literature.
For a  detailed account about lasso and its variants, refer to Table 6 of 
\cite{FFG22} or Fig. 1 of 
\cite{WCY22},  and \cite{ZZ12} 
 and references therein.
\vs
The  penalized regression estimators above improve prediction accuracy meanwhile enhance the interpretability of the model. They, however, pay the price of inducing a little bit of bias in addition to the lack of robustness. There are numerous published articles related to lasso and regularized regression in the literature. However, there are disproportionately few addressing the robustness of the estimators. Are they robust as  supposed (or expected)? Or rather can they resist the influence of just a single contaminated point (or outlier) that is typically buried in high-dimensional data?
\vs
Robust versions of the lasso (or ridge) estimators have been sporadically considered in the literature. The LS in lasso (or ridge), is replaced by M-estimators, as in  \cite{V08} 
 and \cite{LPZ11}; 
or replaced by a Huber-type loss function, as in \cite{RZ04} 
and \cite{SZF20}; 
or by lads, as in \cite{WLJ07}; 
\be
\widehat{\bs{\beta}}_{lad-lasso}=\arg\min_{\bs{\beta}\in \R^p}\Big\{\sum_{i=1}^n |r_i|+\lambda\sum_{i=1}^p|\beta_i| \Big\}, \label{lad-lasso.eqn}
\ee
\noin
or  replacing correlations in lars by a robust type of correlation, 
as in \cite{KVZ07} 
(Rlars); or by S- (\cite{RY84}) 
and MM- (\cite{Y87}) 
estimators, as in \cite{M11} for ridge regression (Rrr); or by the least trimmed squares (LTS) (\cite{R84}), as in 
\cite{ACG13}.
The LTS is defined as
\be
\widehat{\bs{\beta}}^n_{lts}:=\arg\min_{\bs{\beta}\in \R^p} \sum_{i=1}^h (r^2)_{i:n},\label{lts-1.eqn}
\ee
where $(r^2)_{1:n}\leq (r^2)_{2:n}\leq \cdots, (r^2)_{n:n}$ are the ordered squared residuals, $\lceil{n/2}\rceil\leq h\leq n$, and $\lceil \cdot\rceil$ is the ceiling function. \cite{ACG13} replaced the SSR by the objective function of LTS  and defined
\be
\widehat{\bs{\beta}}_{lts-lasso}=\arg\min_{\bs{\beta}\in \R^p}\Big\{ \sum_{i=1}^h (r^2)_{i:n}+ h\lambda\sum_{i=1}^p|\beta_i|\Big\}, \label{lts-lasso.eqn}
\ee
\vs
The idea of \cite{ACG13} has extended to logistic regression with elastic net penalty in 
\cite{KHF18}, and penalized weighted M-type estimators for the logistic regression have also been studied in 
\cite{BBC22}.   

\vs
Most estimators above (except Rlars, Rrr, and $\widehat{\bs{\beta}}_{lts-lasso}$), like both $L_1$ 
 and $L_2$ (LS) estimators, unfortunately, have a pathetic $0\%$ asymptotic breakdown point (i.e., one bad point can ruin (break down) the estimator),
 in sharp contrast to the $50\%$ of the least sum of squares of trimmed (LST) residuals estimator (see Section 3.1 of \cite{ZZ22} or Section 3 here). \cite{KHF18} and \cite{BBC22} both assert their estimators are robust, but no qualitative robustness assessment of their estimators has been established yet. The same situation with the estimator in 
 \cite{SWS22}.
\vs
Now let us take a close look at the three exceptions above.
 The main drawback of the Rlars 
 is the lack of a natural definition or a clear objective function, as commented in \cite{ACG13}.
 The main focus of \cite{M11}
  is robustifying ridge regression (Rrr).
  \vs
   Only $\widehat{\bs{\beta}}_{lts-lasso}$ in \cite{ACG13} has an \emph{established} high finite sample breakdown point (see Section 3 for definition).
    Their  result,  though covers the lasso-type estimators, does not cover the elastic nets and other estimators;
 the authors failed to (i) explain why their estimator can have a breakdown point higher than $50\%$ and (ii) study the properties (such as equivariance and consistency) of their estimator. 
Furthermore, the LTS is notorious for its inefficiency (i.e., usually has a large variance). On the other hand,  the LST introduced in \cite{ZZ22} can outperform the LTS (especially in efficiency) as demonstrated in \cite{ZZ22}. \vs
Based on  observations above, questions we want to address are:
\tb{(i)} Can one replace the LS with a robust LST  in the penalized regression?
How does the resulting estimator perform?
\tb{(ii)} Is it more robust, compared with existing ones? Can one provide a more general breakdown robustness assessment that covers more regularized regression estimators and
 provide an explanation of a breakdown point higher than 50\%?
 \tb{(iii)} Besides robustness, what are other desirable properties for a regression estimator?\vs

The main contributions of this article include (i) it proves that most leading penalized regression estimators can break down by a single adversary contaminating point; (ii) it, hence, introduces a novel and  robust
penalized least squares of depth trimmed regression estimator ($\widehat{\bs{\beta}}_{lst-enet}$) that outperforms leading competitors in the cases considered; and (iii) it proposes an efficient computational algorithm for the estimator and tests for simulated and real high-dimensional data.
\vs
The rest of article is organized as follows. Section \ref{sec.2} establishes a robust result for general regularized regression estimators and reveals that most of leading estimators (including lasso, lars, and enet)
has the worst breakdown point robustness. Section \ref{sec.3} introduces the least sum of squares of (depth) trimmed residuals (LST) regression and studies its robust property. Section \ref{sec.4} introduces a class of penalized regression estimators based on LST and studies their properties including, existence and uniqueness, robustness, and equivariance. Section \ref{sec.5} is devoted to the establishment of the finite sample prediction error bound
and estimator consistency. Section \ref{sec.6} addresses the computation issue of $\widehat{\bs{\beta}}_{lst-enet}$. Section \ref{sec.7} consists of simulation/comparison study and real data application of five competing methods.
 Section \ref{sec.8} ends the article with some concluding discussions. Proofs of main results  are deferred to an Appendix.
\vs
\noin
\section{Robustness of the penalized regression estimators \label{sec.2}}

Are
existing numerous penalized regression methods mentioned above
robust as they are expected or believed? Or rather can they resist the influence of just a
single outlier (or adversary single-point contamination)? We now formally address this question.

\vs
\noin
\tb{A robustness measure}
\vs
In the finite sample practice,
 the most prevailing quantitative measure of the robustness of any  regression or location  estimators  is the
\emph{finite sample breakdown point}, introduced by 
\cite{DH83}.
\medskip

\vs

\noindent
\textbf{Definition} \cite{DH83} ~
The finite sample \emph{replacement breakdown point} (RBP) of a regression estimator $\mb{T}$ at a given sample
$\mb{Z}^{(n)}=\{\bs{Z}_1, \bs{Z}_2,\cdots, \bs{Z}_n\}$, where $\bs{Z}_i:=(\bs{x}_i', y_i)'$, is defined  as
\begin{equation}
\text{RBP}(\mb{T},\mb{Z}^{(n)}) = \min_{1\le m\le n}\bigg\{\frac{m}{n}: \sup_{\mb{Z}_m^{(n)}}\|\mb{T}(\mb{Z}_m^{(n)})- \mb{T}(\mb{Z}^{(n)})\|_2 =\infty\bigg\},
\end{equation}
where $\mb{Z}_m^{(n)}$
stands for an arbitrary contaminated sample by replacing $m$ original sample points in $\mb{Z}^{(n)}$ with arbitrary points in $\R^{p}$ and 
 $\|\bs{x}\|_q=(\sum_{i=1}^n x_i^q)^{1/q}$ is the $\ell_q$-norm for vector $\bs{x} \in \R^n$.
 \hfill\pend
\vs

 Namely, the RBP of an estimator is the minimum replacement fraction that could drive
the estimator beyond any bounds.  It turns out that both $L_1$ (least absolute deviations) and $L_2$ (least squares) estimators have RBP $1/n$ (or $0\%$) 
whereas LST (introduced in Section 3) can have $(\lfloor{n}/{2}\rfloor-p+2)\big/n$ (or $50\%$) (see Section 3), the highest possible asymptotic value for any regression equivariant estimators (see pages 124-125 of
Rousseeuw and Leroy (1987)
\cite{RL87} or Section 3), where $\lfloor \cdot\rfloor$ is the floor function. 
We now present a general RBP result on the penalized regression estimators.
\vskip 3mm

\noin
\tb{A general result on penalized regression estimators}\vs

\noin
\tb{Theorem 2.1} For any given data set $\bs{Z}^{(n)}=\{(\bs{x}'_i, y_i)', i\in\{1,\cdots, n\}\}$ in $\R^p$ ($p>1$), let $\widehat{\bs{\beta}}^*(\lambda_1, \lambda_2,\gamma, \bs{Z}^{(n)})$ be the penalized regression estimator, which minimizes the objective
\be
O(\bs{\beta}, \lambda_1, \lambda_2,\gamma, \bs{Z}^{(n)}):=
\frac{1}{n}\sum_{i=1}^n \mc{L} (r_i)+ g(\bs{\beta}, \lambda_1, \lambda_2, \gamma),
\label{objective-3.eqn}
\ee
where $\lambda_i, \gamma \geq 0$,  
 the combined penalty function $g(\bs{\beta}, \lambda_1, \lambda_2, \gamma) \geq 0$ and  
the loss function $\mc{L}(x)$ is  non-negative, 
 non-decreasing over $(0, \infty)$,
 $\mc{L}(0)=0$
  and
   $\mc{L}(x)\to \infty$  when $x \to \infty$. 
 Then
$$
\mbox{RBP}(\widehat{\bs{\beta}}^*(\lambda_1, \lambda_2,\gamma, \bs{Z}^{(n)}), \bs{Z}^{(n)})={1}/{n}.
$$
\vs
\noin
\tb{Proof:} see the Appendix. \hfill \pend

\vs
\noin
\tb{Remarks 2.1}
\vs
\tb{(i)} Conditions on $\mc{L}(x)$ are relative loose, they hold automatically if $\mc{L}(x)$ is non-negative, non-decreasing, and convex in $x$ and $\mc{L}(0)=0$. The   $\mc{L}(x)$  in theorem covers almost all loss functions in Table 6 of \cite{FFG22}. The penalty function $g(\bs{\beta},\lambda_1,\lambda_2,\gamma)$ covers almost all existing ones  including, among others, $\lambda_1\|\bs{D}_1\bs{\beta}\|^{\gamma}_{\gamma}+\lambda_2\|\bs{D}_2\bs{\beta}\|^2_2$, with  $\bs{D}_i$ being penalty matrices.
\vs
\tb{(ii)} The RBP result in the theorem is very general since the loss function covers most of the existing loss functions in the machine learning and AI literature, e.g., the most popular ones: negative log-likelihood; the $\ell_1$ loss, the $\ell_2$ loss (or any $\ell_q$ loss with $q\geq1$), Huber loss, and the loss of the lasso and most of its variants (see Table 6 of \cite{FFG22}).  The penalty format covers most of the existing ones in the 
literature (indeed, it covers all twenty-five penalty functions listed in Table 1 of \cite{WCY22}).

\vs
\tb{(iii)} The great generality of the result in the theorem implies that most of the existing penalized regression (and \tb{the classic L1 and L2}) estimators are not robust. In fact, they can break down  with just one single outlier (or contaminating point) which often buries in high dimensional data.
\hfill \pend
\vs
Now that most of the existing penalized regression estimators can be broken down by a single outlier (or single-point contamination). Furthermore, existing robust penalized regression estimators are most ad hoc, e.g., Rlars of \cite{KVZ07} is for robustifying lars, and Rrr of \cite{M11} 
 is  for robustifying ridge regression, and  \cite{BBC22} is mainly for robustifying the penalized logistic regression estimators.
\vs
 Only $\widehat{\bs{\beta}}_{lts-lasso}$ of \cite{ACG13} and $\widehat{\bs{\beta}}_{enetLTS}$ of \cite{KHF18} that employed LTS to replace the SSR in lasso have really high breakdown robustness meanwhile do the variable selection job.
But the major drawback of the LTS is its inefficiency (it has a larger variance) as demonstrated in \cite{ZZ22} (also Sections 3 and 7) and Figures 2 and 3 of \cite{KHF18}. \vs
 A natural question is: can one construct a penalized regression estimator that is robust against the outliers or contamination and more efficient (i.e., with a smaller variance than the LTS)?
In the following, we achieve this goal by introducing a robust alternative to the least squares estimator, called an LST (least squares of depth trimmed residuals estimator), and applying it to the penalized regression setting.

\section{The least sum of squares of trimmed residuals regression \label{sec.3}}

\noin
\tb{Definition of LST}

\vs
\noin
To robustify the LS estimator, \cite{R84}
introduced least trimmed squares (LTS) estimator. The procedure orders the squared residuals and then trims the larger ones and keeps at least $ h \geq \lceil n/2\rceil$ squared residuals,
  the  minimizer of the sum of those \emph{trimmed squared residuals} is called an LTS estimator as defined in (\ref{lts-1.eqn}).
$\widehat{\bs{\beta}}^n_{lts}$ is highly robust 
 but is not very efficient, as reported in \cite{MMY06} (page 132) or in \cite{SHH00} 
 having just $7\%$ or $8\%$  asymptotic efficiency. 
A more efficient competitor,  least sum of squares of trimmed (LST) residuals estimator, is
 introduced in \cite{ZZ22}, overcoming LTS drawback while sharing its high robustness and fast computation advantages.
\vs

For a given sample $\mb{Z}^{(n)}=\{(\bs{x}'_i, y_i)', i\in\{1,\cdots, n\}\}$ in $\R^{p}$ and a $\bs{\beta} \in \R^p$, define
$\mu(\bs{Z}^{(n)},\bs{\beta})=\mbox{Med}_i\{r_i\}, ~ 
\sigma(\bs{Z}^{(n)},\bs{\beta})=\mbox{MAD}_i\{r_i\}, 
$
where  $r_i$ is defined in (\ref{residual.eqn}), Med$_i\{r_i\}=\mbox{median}\{r_i, i\in\{1,\cdots, n\}\}$ is the median of $r_i$s, and  MAD$_i\{r_i\}=\mbox{Med}(\{|r_i-\mbox{Med}(r_i)|,~ i\in\{1,\cdots, n\}\})$ is the median of absolute deviations to the center (median) of $r_i$s. Operators Med and MAD are used for discrete data sets (and distributions as well).
\vs
The outlyingness (or equivalently, depth) 
 of a point $x$ in \cite{Z03} is defined to be (strictly speaking, depth=1/(1+outlyingness))
\be
D(x, X^{(n)})=|x-\mbox{Med}(X^{(n)})|/\mbox{MAD}(X^{(n)}),  \label{outlyingness.eqn}
\ee
where $X^{(n)}=\{x_1, \cdots, x_n\}$ is a data set in $\R^1$. 
It is readily seen that $D(x, X^{(n)})$ is a generalized standard deviation, or equivalent to the one-dimensional projection depth/outlyingness (see \cite{ZS00}, \cite{Z03, Z06}
for a high dimensional version). For notion of outlyingness, cf  ~\cite{S81}, and  \cite{DG92}.
For a given $\alpha$ (throughout constant $\alpha\geq 1$, default value is one) in the depth trimming scheme, consider the quantity
\be
Q^n(\bs{\beta}):=Q(\bs{Z}^{(n)}, \bs{\beta}, \alpha)= \frac{1}{n}\sum_{i=1}^{n}r_i^2\mathds{1} \left( D(r_i, R^{(n)})   
\leq \alpha\right), \label{objective-1.eqn}
\ee
 where $\mathds{1}(A)$ is the indicator of $A$ (i.e., it is one if A holds and zero otherwise) and $R^{(n)}:=\{r_i, i \in \{1, 2, \cdots, n\}\}$.
Namely, residuals with their outlyingness (or depth) greater than $\alpha$ (or less than $1/(1+\alpha)$) will be trimmed.
When there is a majority ($\geq \lfloor(n+1)/2\rfloor$) identical $r_i$s, we define $\sigma(\mb{Z}^{(n)}, \bs{\beta})=1$.
Minimizing $Q(\bs{Z}^{(n)}, \bs{\beta}, \alpha)$, one gets the \emph{least} sum of \emph{squares} of {\it trimmed} (LST)  residuals  estimator,
\be
\widehat{\bs{\beta}}^n_{lst}:=\widehat{\bs{\beta}}_{lst}(\mb{Z}^{(n)}, \alpha)=\arg\min_{\bs{\beta}\in \R^p}Q(\bs{Z}^{(n)}, \bs{\beta}, \alpha).\label{lst.eqn}
\ee
Compared with the LTS definition (\ref{lts-1.eqn}), it is readily seen that both estimators trim residuals. However, there are two essential differences:
 \tb{(i)}
the trimming schemes are different. The LTS employs a rank-based trimming scheme that focuses only on the relative position of points (squared residuals) with respect to others and ignores the magnitude of the point and the relative distance between points whereas the LST exactly catches these two important attributes. 
It orders data from a center (the median) outward and trims the points that are far away from the center. This is known as depth-based trimming.
\tb{(ii)} Besides the trimming scheme difference, there is another difference between the LTS and the LST, that is, the order of trimming and squaring. In the LTS, squaring is first, followed by trimming whereas, in the LST, the order is reversed.
\vs
All the difference leads to an unexpected performance difference in the LTS and the LST as demonstrated in the small illustration example in Figure 1 (see Ex 1.1 of \cite{ZZ22}).\vs

\bec
\begin{figure}[h]
    \centering
    \vspace*{-10mm}
    \begin{subfigure}[t]{0.47\textwidth}
    \includegraphics[width=\textwidth]{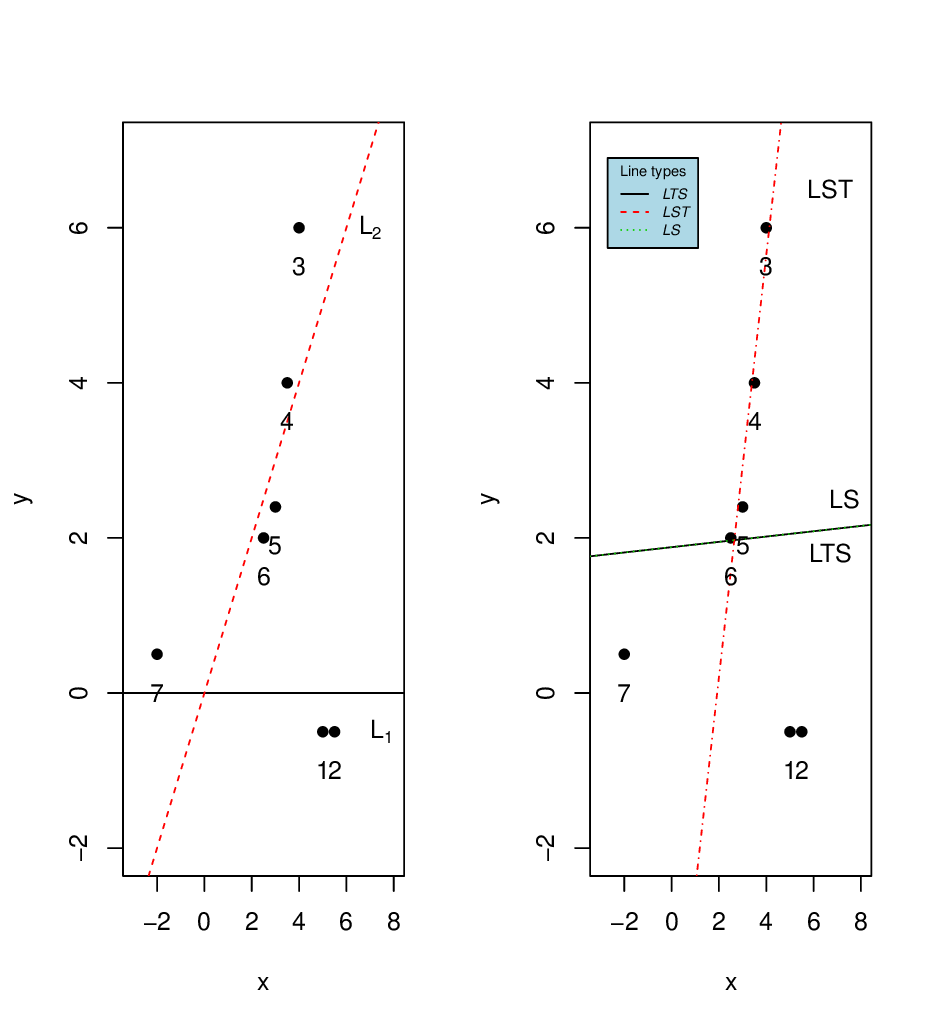}            
        \caption{Left panel: plot of seven artificial points and two candidate lines ($L_1$ and $L_2$), which line would you  pick?
        Sheerly based on the trimming scheme and objective function value, if one uses the number $h=\lfloor n/2\rfloor+\lfloor (p+1)/2\rfloor$ given on page 132 of \cite{RL87}, that is, employing four smallest squared residuals, then the LTS prefers $L_1$ to $L_2$ whereas the LST reverses the preference.\\[1ex] Right panel:  the same seven points are fitted by the LTS, the LST, and the LS (benchmark). A solid black  line is the LTS given by ltsReg. Red dashed line is given by the LST, and green dotted line is given by the LS - which is identical to the LTS line in this case.}
        \label{fig:seven-points-lines}
     \end{subfigure}
     \hspace*{2mm}
    \begin{subfigure}[ht]{0.47\textwidth}
    \vspace*{-15mm}
    \includegraphics[width=\textwidth]{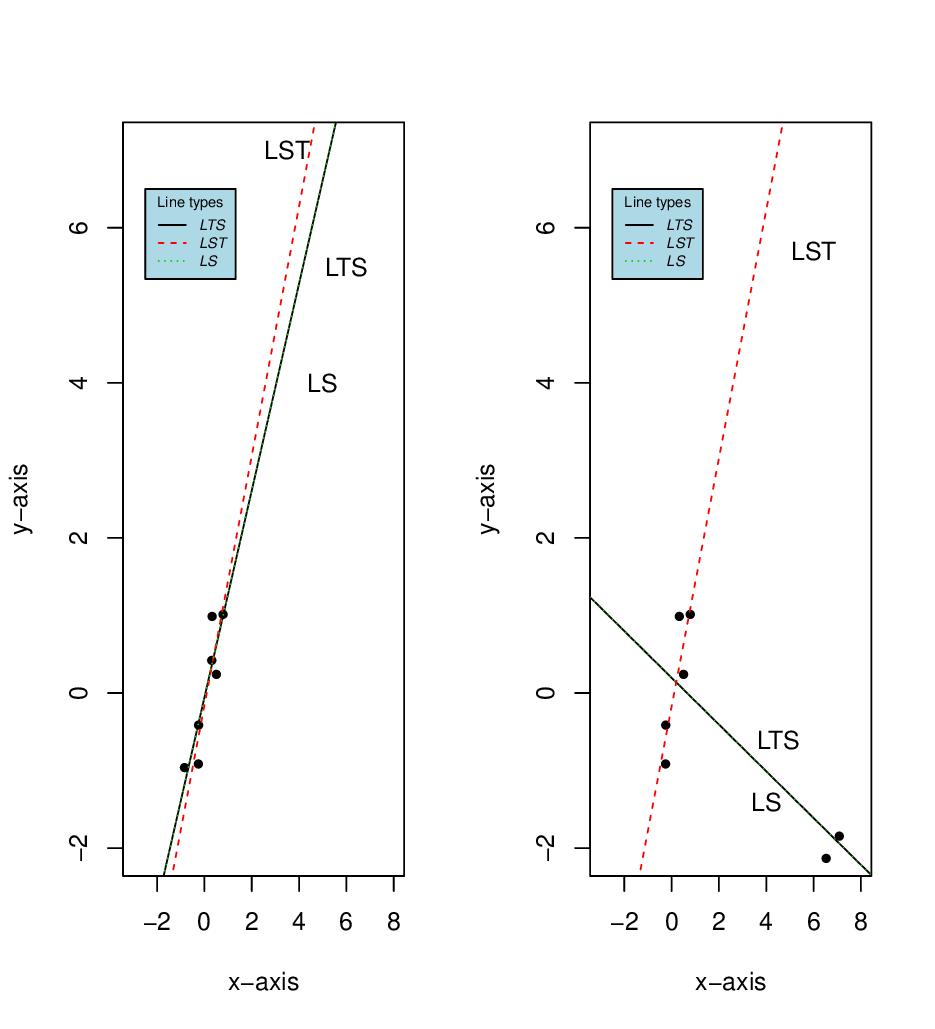}  
        \caption{Left panel: plot of seven highly correlated normal points (with mean being the zero vector and covariance matrix with diagonal entries being one and off-diagonal entries being 0.88) and three lines given by the LST , the LTS, and the LS. The LS line is identical to the LTS line again.\\[1ex] 
        Right panel: the LTS line (solid black) and the LST line (dashed red), and the LS (dotted green) for the same seven highly correlated normal points but with two points contaminated nevertheless. The LS line is identical to the LTS line due to the attributes in the R function ltsReg that is based on 
         \cite{RVD06}).}
        \label{fig:lts-lines}
   \end{subfigure}
   \caption{\footnotesize (a) Performance difference between the LST and the LTS. (b) Performance difference between the LST and the LTS when there are contaminated points ($x$-axis leverage points).}
   \label{Fig:one}
\vspace*{-0mm}
\end{figure}
\enc
\vspace*{-7mm}
Existence and uniqueness of $\widehat{\bs{\beta}}^n_{lst}$ have been addressed in \cite{ZZ22}, it is also equivariant (see \cite{ZZ22}).
A regression estimator $\mb{T}$ is called \emph{regression, scale, and affine equivariant}
 if,  respectively (see page 116 of \cite{RL87}) with $N=\{1,2, \cdots, n\}$
  \bee \mb{T}\left(\{(\bs{w}'_i, y_i+\bs{w}'_i \mb{b})', i\in N\}\right)&=&
\mb{T}\left(\{(\bs{w}'_i, y_i)', i\in N\}\right)+\mb{b}, ~\forall~ \mb{b}\in\R^p \label{regression.equi}\\
\mb{T}\left(\{(\bs{w}'_i, s y_i)', i\in N\}\right)&=&
s\mb{T}\left(\{(\bs{w}'_i, y_i)', i\in N\}\right), ~\forall~ s\in\R^1\\
\mb{T}\left(\{(A'\bs{w}_i)', y_i)', i\in N\}\right)&=&
A^{-1}\mb{T}\left(\{(\bs{w}'_i, y_i)', i\in N\}\right),~\forall~ \text{nonsingular}~ A\in \R^{ p\times p}.
\ene
\vs
Now with the measure of robustness (presented in the last section), naturally one wants to ask the question:
is $\widehat{\bs{\beta}}^n_{lst}$ theoretically more robust than the LS estimator $\widehat{\bs{\beta}}^n_{ls}$? 

 \vs
\noindent
\tb{Robustness of LST}
\vs
We shall say  $\mb{Z}^{(n)}$  is\emph{ in general position}
when any $p$ of observations in $\mb{Z}^{(n)}$ gives a unique determination of $\bs{\beta}$.
In other words, any (p-1) dimensional subspace of the space $(\bs{x'}, y)'$ contains at most p observations of
$\mb{Z}^{(n)}$.
When the observations come from continuous distributions, the event ($\mb{Z}^{(n)}$ being in general position) happens with probability one. \vs

\noin
\tb{Theorem 3.2} \cite{ZZ22} For $\widehat{\bs{\beta}}^n_{lst}$ defined in (\ref{lst.eqn}) and $\mb{Z}^{(n)}$ in general position, we have 
\be
\text{RBP}(\widehat{\bs{\beta}}^n_{lst}, \mb{Z}^{(n)})=\left\{
\begin{array}{ll}
\lfloor (n+1)/2\rfloor\big/n, & \text{if $p=1$,}\\[1ex]
(\lfloor{n}/{2}\rfloor-p+2)\big/n,& \text{if $p>1$.}\\
\end{array}
\right. \label{T*-bp.eqn}
\ee
\hfill \pend

\noin
The LST not only shares the best $50\%$ asymptotic breakdown value of the LTS, it is much more efficient than the LTS as demonstrated in the Table 1 below (see \cite{ZZ22}).\vs
\begin{table}[!h]
\centering
~~ Normal data sets, each with $\varepsilon \%$ contamination\\
~~ Table entries (a, b) are: a:=empirical mean squared error,  b:=total time consumed  
\bec
\begin{tabular}{c c c c c c }
~ & ~& ~~~~~~~~~~~~~$\varepsilon=5\%$ &&~~~~~~~~~~~~$\varepsilon=10\%$ &\\
~p~ & ~n~& ~~AA1~~  &~~ltsReg~~& AA1~~&ltsReg \\
\hline\\[0.ex]
 & 100& (0.2971,~~9.6581)& (0.3010,~~22.867)& (0.2843,~~494.01)& (0.2942,~~ 25.289)\\
5&200&(0.2503,~~26.045)& (0.2650,~~41.861)& (0.2517,~~26.629)&(0.2630,~~ 43.504)\\
& 300&(0.2396,~~54.100)& (0.2551,~~63.639)& (0.2366,~~54.885)&(0.2534,~~ 63.522)\\[2ex]
 & 400& (0.1335,~~1085.6)& (0.1394,~~181.18)& (0.1340,~~1056.2)&(0.1382,~~175.92)\\
10&500&(0.1280,~~1207.7)& (0.1321,~~222.81)& (0.1289,~~1178.5)&(0.1321,~~218.94)\\
& 600&(0.1247,~~1308.4)& (0.1285,~~152.47)& (0.1253,~~1273.6)&(0.1276,~~149.99)\\[2ex]
 & 700& (0.0815,~~2044.9)& (0.0885,~~549.61)& (0.0838,~~1994.0)&(0.0882,~~547.53)\\
20&800&(0.0776,~~2261.7)& (0.0837,~~620.63)& (0.0796,~~2177.0)&(0.0837,~~616.87)\\
& 900&(0.0748,~~2436.1)& (0.0804,~~541.20)& (0.0761,~~2353.7)& (0.0795,~~ 538.43)\\[2ex]
~ & ~& ~~~~~~~~~~~~~$\varepsilon=30\%$ &&~~~~~~~~~~~~$\varepsilon=40\%$ &\\
&300&(0.4347,~~53.248)& (1.9236,~~1635.1)& (0.4352,~~56.430)& (1.3517,~~1712.8)\\
40&400&(0.3362,~~100.04)& (1.2604,~~2401.5)& (0.3314,~~102.81)&(0.8995,~~2399.5)\\
&500&(0.2594,~~147.66)& (0.9514,~~ 2963.4)& (0.2873,~~146.67)&(0.6851,~~2787.7)\\[2ex]
&300&(0.5242,~~58.736)& (2.7826,~~2861.8)& (0.5700,~~59.903)&(1.9808,~~2896.3)\\
50&400&(0.4085,~~ 89.897) &(1.7562,~~3292.0)& (0.4539,~~108.88)&(1.2547,~~3925.5)\\
&500&(0.3107,~~145.84)& (1.2870,~~4510.5)& (0.3406,~~145.75)&(0.9086,~~4419.6)\\[0ex]
\hline
\end{tabular}
\enc
\caption{\footnotesize Total computation time  for all $1000$ samples (seconds) and empirical mean squared error (EMSE, see definition in (\ref{emse.eqn})) of the LST (AA1) versus the LTS (ltsReg) for various $n$s, $p$s, and contaminations. AA1 stands for the algorithm to compute the LST.}
\label{lts-vs-lst-alpha-30}
\end{table}

Inspecting the table reveals that (i) in terms of empirical mean squared error (EMSE), AA1 (or rather the LST) is the overall winner (with the smallest EMSE in all cases considered), the LTS has the largest EMSE in all the cases;
 (ii) in terms of speed, the LTS (or rather ltsReg) is the winner when $p=10$ or $20$. AA1 is the winner for all other $p$'s, except when $p=5$, $n=100$ and $\varepsilon=10\%$. For the latter case, AA1 can still be the faster if tuning a parameter in AA1, 
 then one gets $(0.2986,10.396)$ for AA1 versus $(0.2948,23.133)$ for ltsReg. 
\vs
\section{A  class of penalized regression estimators based on the LST \label{sec.4}} 

\noin
\tb{Definition}~~
\vs
Now that we have a much more robust regression estimator than the LS, which turns out to be more efficient than the LTS.
It is quite natural to replace the SSR in (\ref{elastic-nets.eqn})
 by the $Q^n$ defined by (\ref{objective-1.eqn}), and minimize it, subject to two constraints: $\ell_{\gamma}$-constraint $\sum_{i=1}^p|\beta_i|^{\gamma}\leq t_1$, $t_1\geq 0$, $\gamma \geq 1$; and $\ell_2$-constraint $\sum_{i=1}^p\beta^2_i \leq t_2$, $t_2\geq 0$,
  the minimizer  is 
\be
\widehat{\bs{\beta}}^n_{lts-enet} (\alpha, \lambda_1, \lambda_2, \gamma):=\arg\min_{\bs{\beta}\in \R^p}\Big\{\frac{1}{n}\sum_{i=1}^{n}r_i^2w_i+\lambda_1 \sum_{i=1}^p|\beta_j|^{\gamma} +\lambda_2\sum_{i=1}^p\beta^2_i\Big\}, \label{lst-en.eqn}
\ee
where $\lambda_i:=\lambda(t_i)\geq 0$, $\alpha,~\gamma\geq 1$, and $w_i:=w_i(\bs{\beta}):=w_i(\bs{\beta}, r_i, \bs{Z}^{(n)})=\mathds{1}\left( D(r_i, R^{(n)})\leq \alpha \right)$.
\vs

Before studying its robustness, we address existence and uniqueness of $\widehat{\bs{\beta}}^n_{lst-enet}(\alpha,\lambda_1,\lambda_2,\gamma)$.
\vs
\vs
\noin
\tb{Existence and uniqueness}
\vs
Existence and uniqueness are implicitly assumed for many other penalized regression estimators in the literature. We formally address them below for $\widehat{\bs{\beta}}^n_{lst-enet}(\alpha,\lambda_1,\lambda_2,\gamma)$.
\vs
\vs
\noindent
\tb{Theorem 4.1}
\bi
\item[(i)]
$\widehat{\bs{\beta}}^n_{lst-enet}(\alpha, \lambda_1, \lambda_2, \gamma)$ in (\ref{lst-en.eqn})  always exists; 
\item[(ii)]  $\widehat{\bs{\beta}}^n_{lst-enet}(\alpha, \lambda_1, \lambda_2, \gamma)$ in (\ref{lst-en.eqn}) is unique provided that (a) $\lambda_1>0$ and $\gamma>1$ or (b) $\lambda_2>0$.
\ei
\vs

\noindent
\tb{Proof}: see the Appendix. \hfill \pend
\vs
The proof of above theorem needs the following result.
\vs
 \noindent
 \tb{Lemma 4.1} Let $S\subset \R^p$ be an open set and $f(\bs{x})$: $\R^p \to \R^1$ be  strictly convex over $S$ and continuous over $\overline{S}$ (the closure of $S$). Let $\bs{x}^*$ be the global minimum of $f(\bs{x})$ over $S$ and $\bs{y}$ be a point on the boundary of $S$,  then $f(\bs{y}) > f(\bs{x}^*)$.
 \vs
 \noin
 \tb{Proof:} see the Appendix. \hfill \pend
 \vs

 \vs
\noin
\tb{Remarks 4.1}\vs
\bi
\item[(i)] Note that $\widehat{\bs{\beta}}^n_{lst-enet}(\alpha, 0,0, \gamma)=\widehat{\bs{\beta}}^n_{lst}$.
 A sufficient condition for its uniqueness is $\bs{C}_n:=\bs{X}'_n\mbox{diag}(w_i, \cdots, w_n)\bs{X}_n/n$ being invertible. That is,  the rank of  $\bs{X}_n$ and the matrix formed by any its $k:=\sum_iw_i$ sub-rows is $p$ (see \cite{ZZ22}). However, in many applied data set cases, the number of variables ($p$) is even larger than the number of observations ($n$), we must have rank $< p$. So it might not be unique. However, if (a) or (b) in (ii) of the theorem holds, then the strictly convexity guarantees the uniqueness of $\widehat{\bs{\beta}}^n_{lts-enet}(\alpha, \lambda_1, \lambda_2, \gamma)$.

\item[(ii)] The uniqueness of $\widehat{\bs{\beta}}^n_{lts-enet}(\infty, \lambda_1, 0, \gamma)$ (here $\alpha=\infty$ is in the sense that $\alpha \to \infty$), that is, the uniqueness of $\widehat{\bs{\beta}}_{bridge} (\lambda_1, \gamma)$ has been intensively discussed in the literature, see e.g.,
in the Theorems 1 and 2 of \cite{F98}, 
it was shown that $\widehat{\bs{\beta}}^n_{lts-enet}(\infty, \lambda_1, 0, \gamma)$ is unique if $\lambda_1>0$ and $\gamma> 1$ plus  some condition on the Hessian matrix of SSR;
 in their Lemma 2 ($\gamma=1$), \cite{ZH05} 
  showed that it is not unique when there is repeated row of $\bs{X}_n$,  \cite{TR13} 
   and  \cite{AT18} 
    ($\gamma=1$) argued that it is unique with probability one under the some assumption on predictor variables. Also see section 2.6 of \cite{HTW15}.
    \hfill \pend
 \ei
\vs
\noin
The most relevant question now is: Is $\widehat{\bs{\beta}}^n_{lst-enet}(\alpha,\lambda_1,\lambda_2,\gamma)$ much more robust than the existing ones?  Or rather, what is its RBP? 
The next result covers both the LST and the LTS based regularized estimators and provides an affirmative answer to the question.
\vs
\noindent
\tb{Theorem 4.2}  Let $\widehat{\bs{\beta}} (\lambda_1,\lambda_2,\gamma,\bs{Z}^{(n)})$ be the penalized regression estimator which minimizes the objective function
\be
Q(\bs{\beta}, \lambda_1, \lambda_2, \gamma, Z^{(n)}):=\frac{1}{n}\sum_{i=1}^n r_i^2 w_i+\lambda_1\sum_{i=1}^p|\beta_i|^{\gamma}+\lambda_2\sum_{i=1}^p\beta^2_i, \label{objective-45.eqn}
\ee
where $w_i\in\{0, 1\}$ is an indicator function: $\mathds{1}(r_i^2\leq r^2_{h:n})$ or $\mathds{1}(D(r_i, R^{(n)})\leq \alpha)$ and $\sum_{i=1}^nw_i=k$ $(\lceil n/2\rceil\leq k \leq n )$, $\lambda_i\geq 0$, and $\lambda_1+\lambda_2>0$,
$1\leq \gamma\leq 2$. Then
$$
\mbox{RBP}(\widehat{\bs{\beta}} (\lambda_1,\lambda_2,\gamma,\bs{Z}^{(n)}), \bs{Z}^{(n)})={(n-k+1)}/{n}.
$$
\vs
\noindent
\tb{Proof}: see the Appendix. \hfill \pend
\vs
\noin
\tb{Remarks 4.2}
\vs
\tb{(i)} The square loss function in the theorem (or in (4.20)) 
can be easily extended to a more general $\mc{L}$ such as the one defined in Theorem 3.1,
the RBP result still holds. The result thus covers the main result (Theorem 1) of  \cite{ACG13}, where $k=h$ (the default value is $\lfloor(n+p+1)/2 \rfloor$). Indeed, to achieve better robustness one has to trim some squared residuals.
 The theorem covers the RBP of $\widehat{\bs{\beta}}^n_{lst-enet}(\alpha,\lambda_1,\lambda_2,\gamma)$ in (4.19) 
 for any $\alpha\geq 1$, which reaches its highest value $(\lfloor n/2\rfloor+1)/n$ when $\alpha=1$ ($k=\lfloor (n+1)/2 \rfloor$ in this case). It also first time provides the RBP for the reweighted sparse-LTS (or enet-LTS ) estimators in
\cite{ACG13} (or \cite{KHF18}) with $k=n_w$ there. The theorem tells that Ridge, Bridge
, lasso, and enet all have the lowest RBP $1/n$.\vs
\tb{(ii)}
Notice that the RBP result is dimension-free, it is even higher than the upper bound for any \emph{regression equivariant} estimator (see Theorem 4 on page 125 of \cite{RL87}).
The main reason for this is that the estimator violates the regression equivariance.
\vs
\tb{(iii)}  Without the regression equivariance, any constant vector will have the best possible RBP ($100\%$), but it is not a good estimator at all. Note that the RBP definition in \cite{ACG13} and  
 \cite{KBW18} is different from the traditional one. Furthermore, theorem 2 and remark 2 on RBP in \cite{KBW18} are debatable.
\hfill \pend
\vs \vs
\noin
\tb{Equivariance}
\vs
Among \emph{regression, scale, and affine equivariance}, the three desired properties (discussed in \ref{sec.3}), the regression equivariance is the most fundamental, it demands that if one shifts response variable $y$ up and down, then the regression line (or hyperplane) should shift accordingly up and down. The LS estimator and all its robust alternatives mentioned so far satisfy the three properties. But this is not the case for most of regularized regression estimators. In fact,
\vs
\noindent
\tb{Theorem 4.3} ~~ 
 Among three equivariant properties, only scale equivariance is processed by  $\widehat{\bs{\beta}}_{ridge}$ in (\ref{ridge.eqn}), the $\widehat{\bs{\beta}}_{sqrt-lasso}$ in (\ref{squre-rrot-lasso.eqn}), the $\widehat{\bs{\beta}}_{lad-lasso}$ in (\ref{lad-lasso.eqn}), and $\widehat{\bs{\beta}}^n_{lts-enet}(\alpha, 0, \lambda_2, \gamma)$ in (\ref{lst-en.eqn}) among all penalized
regression estimators discussed previously.
\vs
\noindent
\tb{Proof:} scale equivariance of the $\widehat{\bs{\beta}}_{ridge}$, the $\widehat{\bs{\beta}}_{sqrt-lasso}$,  the $\widehat{\bs{\beta}}_{lad-lasso}$ , the $\widehat{\bs{\beta}}_{lad-lasso}$,  and $\widehat{\bs{\beta}}^n_{lts-enet}(\alpha, 0, \lambda_2, \gamma)$  is trivial verification in light of (\ref{ridge.eqn}), (\ref{squre-rrot-lasso.eqn}), (\ref{lad-lasso.eqn}), and (\ref{lst-en.eqn}). For other properties and penalized estimators, it suffices to show that regression equivariance is violated.\vs
When $y_i$ is shifted to $y_i+\bs{w}'_i \mb{b}$, if the regression coefficients $\bs{\beta}$ is also shifted to $\bs{\beta}+\bs{b}$, then SSR is unchanged whereas the constraint or penalty still on $\bs{\beta}$.
 \hfill \pend
\vs
\noindent
\tb{Remarks 4.3}
\vs
\tb{(i)}
There has been an abundance of theoretical and computational work on the generalized
lasso and its variants and its special cases. Among hundreds, if not thousands, publications on
penalized regression in the literature, very few addressed equivariance. Exceptions are  \cite{OAC16},
\cite{KBW18}, and \cite{SO11}. 
\cite{OAC16} admitted
that their shooting S-estimator fails to meet the regression equivariance. \cite{KBW18} asserted
that via transformation and re-transformation, their estimator enjoys the three equivariance
properties, which, however, is debatable. \cite{SO11} asserted that their IPOD estimate
$\widehat{\bs{\beta}}$ processes the three desired equivariant properties.  \vs
\tb{(ii)} Standardizing $y$ and $\bs{x}$ columns are common practice in the literature for many computational algorithms for regularized estimators. This, however, amounts to assuming implicitly that
these estimators meet the three equivariance properties. Furthermore, centering the observations of $y$ and $\bs{x}$ might spread the contamination or outlyingness.
\hfill \pend
\vs
\section{Finite sample predition error bounds--consistency \label{sec.5}}
\noin
In this section we assume that the true model is $\bs{Y}=\bs{X}\bs{\beta}_0+\bs{e}$ where $\bs{Y}=(y_i,\cdots, y_n)'$, $\bs{X}=(\bs{w}_1,\cdots, \bs{w}_n)'$, and $\bs{e}=(e_1,\cdots, e_n)'$ with $y_i$, $e_i$, and $\bs{w}_i$ defined in (\ref{model-population.eqn}) and  (\ref{residual.eqn}).
  We investigate the difference between $\bs{X}\widehat{\bs{\beta}}^n_{lst-enet}$ and $\bs{X}\bs{\beta}_0$ (prediction error). Write $\widehat{\bs{\beta}}^n$ for $\widehat{\bs{\beta}}^n_{lst-enet}$ for simplicity. \vs 
  Define an index set $I(\bs{\beta}):=\{i: w_i=1\}$, 
  the scalar $w_i\in \{0, 1\}$ in (\ref{lst-en.eqn}) is different from the vector $\bs{w}_i$ above. 
  Write ${D}(\bs{\beta})=\mbox{diag}(w_1, \cdots, w_n)$ with $w_i$ defined in (\ref{lst-en.eqn}). Let ${A}$ be a $n$ by $n$ symmetric positive semidefinite  matrix, a norm (or seminorm) induced by ${A}$ is $\|\bs{x}\|^2_{{A}}=\bs{x}'{A}\bs{x}$ for any $\bs{x}\in \R^n$. Although $\widehat{\bs{\beta}}^n$ provides predictions for all $i$, but we just employed residuals $r_i$ with $i\in I(\widehat{\bs{\beta}}^n)$ in (\ref{lst-en.eqn}), so instead of
  looking at $\|\bs{X}\big(\widehat{\bs{\beta}}^n-\bs{\beta}_0\big)\|^2$, we will focus on the squared perdition error $\|\bs{X}\big(\widehat{\bs{\beta}}^n-\bs{\beta}_0\big)\|^2_{D(\widehat{\bs{\beta}}^n)}$.
\vs
\noindent
\tb{Lemma 5.1} Assume that $\bs{\beta}_0$ is the true parameter of the model in (\ref{model-population.eqn}), $\widehat{\bs{\beta}}^n:= \widehat{\bs{\beta}}^n_{lst-enet}$ is defined in (\ref{lst-en.eqn}). We have
\begin{align}
\|\bs{X}\big(\widehat{\bs{\beta}}^n-\bs{\beta}_0\big)\|^2_{{D}(\widehat{\bs{\beta}}^n)}&\leq 
 \frac{2}{n} \bs{e}'{D}(\widehat{\bs{\beta}}^n)\bs{X}(\widehat{\bs{\beta}}^n-\bs{\beta}_0)
+\frac{1}{n}\big(\|\bs{e}\|^2_{{D}(\bs{\beta}_0)}-\|\bs{e}\|^2_{{D}(\widehat{\bs{\beta}}^n)}\big) \nonumber\\[1ex]
&+\lambda_1\|\bs{\beta}_0\|^{\gamma}_{\gamma}+\lambda_2\|\bs{\beta}_0\|^2_2-\lambda_1\|\widehat{\bs{\beta}}^n\|^{\gamma}_{\gamma}-\lambda_2\|\widehat{\bs{\beta}}^n\|^2_2. \label{base-inequality.eqn}
\end{align}

\noindent
\tb{Proof}: see the Appendix.
\hfill \pend
\vs
Write $(e^*_1,\cdots, e^*_n):=(\bs{e}^*)'$ with $e^*_i=e_i*\mathds{1}\big(i\in I(\widehat{\bs{\beta}}^n)\big)$. Define two sets
$$
\mathscr{S}_1: = \left\{\max_{1\leq j\leq p}2|(\bs{e}^*)'\bs{x}^{(j)}|/n\leq q_1 \right\}, ~~~
\mathscr{S}_2: = \left\{\|\bs{e}\|^2_{D^*}/\sigma^2-N_d\leq q_2\right\},
$$
where 
$\bs{x}^{(j)}$ is the $j$th column of the fixed  design matrix $\bs{X}_{n \times p}$, $D^*=D(\bs{\beta}_0)-D(\widehat{\bs{\beta}}^n)$, a diagonal matrix with $D^*(i, i)=\mathds{1}(D(\bs{\beta}_0)(i, i)=1 ~\mbox{and}~ D(\widehat{\bs{\beta}}^n)(i,i)=0 )$. Let $N_d=|I(\bs{\beta}_0)|-|I(\bs{\beta}_0)\cap I(\widehat{\bs{\beta}}^n)|$, 
it is readily seen that $0\leq N_d\leq (n-1)$. 
Note that $
\bs{e}'{D}(\widehat{\bs{\beta}}^n)=(e^*_1,\cdots, e^*_n):=(\bs{e}^*)'$. \tb{Assume hereafter that $\bs{\max_{1\leq j\leq p}\|\bs{x}^{(j)}\|_2\leq c_x}$ for a constant $c_x$}.
\vs
In the classical setting
$e_i$ in (\ref{model-population.eqn}) is assumed  $N(0,\sigma^2)$, it is needed for the second result below, but for the first, it can be relaxed to be a sub-Gaussian variable. For the definition of the latter, we refer to  Definition 1.2 of 
\cite {RH17} and/or  Theorem 2.1.1 of 
\cite{P20}.
\vs
\noin
\tb{Lemma 5.2} (i) Let $e_i$s in (\ref{model-population.eqn}) be independent sub-Gaussian variables  that have variance proxy $\sigma^2$, 
then $(\bs{e}^*)' \bs{x}^{(j)}/c_x$ is a sub-Gaussian variable with variance proxy $\sigma^2$. 
(ii) Let $e_i$ in (\ref{model-population.eqn}) be i.i.d. $N(0,\sigma^2)$, then
$\|\bs{e}\|^2_{D^*}/\sigma^2$ follows a ${\chi}^2$ distribution with $N_d$ degrees of freedom.
\vs

\noindent
\tb{Proof}: see the Appendix.
\hfill \pend
\vs

\noin
\tb{Lemma 5.3} Assume that $e_i$s in (\ref{model-population.eqn}) are i.i.d. $N(0,\sigma^2)$ and other assumptions in Lemmas 5.1-5.2 hold, for any $\delta\in (0,1)$ let $$q_1=\frac{4{c_x}\sigma}{n}\Big(2\sqrt{p}+\sqrt{2\log(2/\delta)}\Big);~ \\ 
q_2=2 \sqrt{\log(2/\delta)}\Big(\sqrt{|I(\bs{\beta}_0)|}+\sqrt{\log(2/\delta)}\Big),$$

 then
\be
P(\mathscr{S}_1)\geq 1-\delta/2;  ~~    
P(\mathscr{S}_2)\geq 1-\delta/2.
\ee
\vs
\noindent
\tb{Proof}: see the Appendix. \hfill \pend

\vs
\noin
In light of all Lemmas we are in the position to present the main result.
\vs
\noindent
\tb{Theorem 5.1} Set $\gamma$ in (\ref{lst-en.eqn}) to be one and assume that the assumptions in Lemma 5.3 hold.  For any $\delta\in (0,1)$, selecting $\lambda_1\geq q_1$. Then with probability at least $1-\delta$, one has
\be
\|\bs{X}(\widehat{\bs{\beta}}^n-\bs{\beta}_0)\|^2_{D(\widehat{\bs{\beta}}^n)}\leq 2\lambda_1\sqrt{p}\|\bs{\beta}_0\|_2+\lambda_2\|\bs{\beta}_0\|^2_2+\frac{\sigma}{n}(q_2+N_d). \label{error-bound.eqn}
\ee
\vs
\noindent
\tb{Remarks 5.1}
\vs
\tb{(i)} If select $\lambda_1 \geq q_1$  and 
 in the order of $O(\sqrt{p}/n$) and $\sqrt{\lambda_2}\leq \lambda_1\sqrt{p}$ and if $\|\bs{\beta}_0\|_2$ is in the order less than $O(n/p)$ (e.g., $o(n/p)$), then one obtains the consistency if $N_d=o(n)$ since $q_2=O(n^{1/2})$.\vs
Theorem 5.1 and  (5.23) 
certainly provide a finite sample squared perdition bound, but
 the assumption of $N_d=o(n)$ above is too arbitrary, it can be dropped nevertheless. Set $\alpha$ in  (4.19) 
 to be one, then $K:=|I(\bs{\beta}_0)|=|I(\widehat{\bs{\beta}}^n)|=
\lfloor(n+1)/2\rfloor$. Treat the two parts of $D^*$ separately, notice that both $\|e\|^2_{D(\bs{\beta}_0)}/\sigma^2$ and  $\|e\|^2_{D(\widehat{\bs{\beta}}^n)}/\sigma^2$ have a $\chi^2$ distribution with the same degrees of freedom $K$. Write $\frac{1}{n}\big(\|\bs{e}\|^2_{{D}(\bs{\beta}_0)}-\|\bs{e}\|^2_{{D}(\widehat{\bs{\beta}}^n)}\big) $ in (5.21)
 as $\frac{\sigma^2}{n}\Big((\|\bs{e}\|^2_{{D}(\bs{\beta}_0)}/\sigma^2-K)+(K-\|\bs{e}\|^2_{{D}(\widehat{\bs{\beta}}^n)}/\sigma^2)\Big) $. Apply the exponential tail bounds on page 1325 of  \cite{LM00},
the upper bound in RHS of (5.23) 
 becomes $2\lambda_1\sqrt{p}\|\bs{\beta}_0\|_2+\lambda_2\|\bs{\beta}_0\|^2_2+2\sigma^2 q_2/n$. The consistency is obtained without $N_d=o(n)$ assumption.
\vs

\tb{(ii)} In above discussions, we treat the unknown $\sigma$ as known. It appears in $q_1$ and in the upper bound of (5.23). 
 In practice, we have to estimate it by an estimator, say $\widehat{\sigma}$ so that
$P(\widehat{\sigma}\geq\sigma)$ with high probability (say, $1-\delta/3$, in this case, if we change $\delta/2$ and $\log(2/\delta)$ in Lemma 5.3 to $\delta/3$ and $\log(3/\delta)$ respectively, then Theorem 5.1 still holds). Such an estimator $\widehat{\sigma}$  has been given on page 104 of 
\cite{BVDG11}.
\vs
\tb{(iii)} One limitation of Theorem 5.1 is that the design matrix is fixed. For the general random design $X$ case, one can treat it following the approaches of  \cite{BMN12} 
and  \cite{GMPT07}. 
\hfill \pend
\vs
\section{Computation algorithm \label{sec.6}}
\vs
\noindent
\tb{Re-parametrizations}\vs
 \tb{(i)}~
Following the notation used in (\ref{objective.eqn}), we
note that the objective function on the RHS of (\ref{lst-en.eqn}) can be written as (also see the proof of Lemma 5.1)
$$O_n(\bs{\beta}, \lambda_1, \lambda_2, \gamma)=
\frac{1}{n}\|\bs{Y}-\bs{X}\bs{\beta}\|^2_{D(\bs{\beta})}+\lambda_1\sum_{j=1}^p|{\beta}_j|^{\gamma}+\lambda_2\|\bs{\beta}\|^2_2,
$$
where ${D}(\bs{\beta})=\mbox{diag}(w_1, \cdots, w_n)$ with $w_i$ defined in (\ref{lst-en.eqn}).
Now for every $\lambda_2>0$, if we write $\bs{X}^*_{(n+p)\times p}=(1+\lambda_2)^{-1/2}(\bs{X}'_{n\times p}, \sqrt{\lambda_2}\bs{I}_{p \times p})'$, $Y^*_{(n+p)\times 1}= (\bs{Y}'_{n \times 1}, \bs{0}'_{p\times 1})'$, $\bs{\beta}^*_{p\times 1}= (1+\lambda_2)^{1/2}\bs{\beta}$. Then $D^*(\bs{\beta}^*)_{(n+p)\times p}:=(D(\bs{\beta^*}), \bs{I}_{p \times p})'=(D(\bs{\beta}), \bs{I}_{p \times p})'$. If let $\lambda^*_1:=\lambda_1/(1+\lambda_2)^{1/2}$, we have
\[
O_n(\bs{\beta}^*, \lambda_1, \lambda_2, \gamma)=O_n(\bs{\beta}, \lambda_1, \lambda_2, \gamma)=\frac{1}{n}\|\bs{Y}^*-\bs{X}^*\bs{\beta}^*\|^2_{D^*(\bs{\beta}^*)}+\lambda_1^*\sum_{j=1}^p|{\beta}^*_j|^{\gamma},
\]
An $\ell_1$-type penalized regression with an objective function much resembling that of a lasso-type problem (especially when in the $\gamma=1$ case). Denote the minimizer of the objective function above by $\widehat{\bs{\beta}}^*$. It can be computed via the approach for lasso such as the lars algorithm of \cite{EHJT04}. 
\vs
\tb{(ii)}~
Alternatively, if we set $\lambda^*=\lambda_1+\lambda_2$  and $\alpha^*=\lambda_2/(\lambda_1+\lambda_2)$ (note that $\lambda_1+\lambda_2>0$, otherwise we have a non-penalized problem addressed in \cite{ZZ22}),
then we have
\be
O_n(\bs{\beta}, \lambda_1, \lambda_2, \gamma)=O_n(\bs{\beta}, \alpha^*, \lambda^*, \gamma):=
\frac{1}{n}\|\bs{Y}-\bs{X}\bs{\beta}\|^2_{D(\bs{\beta})}+\lambda^*\Big((1-\alpha^*)\sum_{j=1}^p|{\beta}_j|^{\gamma}+\alpha^*\|\bs{\beta}\|^2_2\Big). \label{new-objective.eqn}
\ee
Note that $\alpha^*\in [0, 1)$ (a pure ridge regression case is excluded) and $\lambda^*\in (0, \lambda_0]$ for some $\lambda_0$ (which is set to be $\max_{1\leq j\leq p}|2\bs{Y}'\bs{x}^{(j)}|/n$ as in the literature, see e.g., \cite{ACG13} and Section 2.12 of \cite{BVDG11}). {\it Boundedness of parameters} is the advantage of this formulation.
For 
a given data set $\bs{Z}^{(n)}=\{(\bs{x}'_i, y_i)', i\in \{1,\cdots, n\}\}$, we now present the outline our approximate algorithm (AA) for $\widehat{\bs{\beta}}^n_{lst-enet}$.\vs
\noin
\tb{Pseudocode for computing $\widehat{\bs{\beta}}^n_{lst-enet}$ (lst-enet)}
\vs
\tb{(1)} Sample two indices  $\{i, j\}$ (two points) and obtain at least $p$ 
 $\bs{\beta}$s: $\bs{\beta}^k$  ($ k\in\{1,\cdots, p\}$) 
 using the algorithm AA1 in \cite{ZZ22} and obtain index sets $I(\bs{\beta}^k):=\{i: w_i:=w_i(\bs{\beta}^k)=1\}$.
\vs
\tb{(2)} For each $\bs{\beta}^k$, employing the strategy below select a pair $(\alpha^*, \lambda^*)$ with respect to sub-data sets $({D}(\bs{\beta}^k)\bs{X}, {D}(\bs{\beta}^k)\bs{Y})$.\vs
\tb{(3)} Based on the sub-data sets $({D}(\bs{\beta}^k)\bs{X}, {D}(\bs{\beta}^k)\bs{Y})$ obtain  solution $\widehat{\bs{\beta}}^k$ via LARS algorithm (limited the total steps to $900$)
\vs
\tb{(4)} Evaluate $O_n(\bs{\beta}, \alpha^*, \lambda^*, \gamma)$ with respect to ${\bs{\beta}}^k$ and $\widehat{\bs{\beta}}^k$ ($ k\in\{1,\cdots, p\}$). Update $\widehat{\bs{\beta}}^n_{lst-enet}$ (initially it is a $\bs{0}$ vector) to be the one that 
 minimizes
\[O_n(\bs{\beta}, \alpha^*, \lambda^*, \gamma)=\frac{1}{n}\sum_{i\in I(\bs{\beta})}w_ir^2_i(\bs{\beta})+\lambda^*\Big((1-\alpha^*)\sum_{i=1}^p|{\beta}_j|^{\gamma}+\alpha^*\|\bs{\beta}\|^2_2\Big).
\]
\vs
\tb{(5)} Repeat \tb{(1)-(4)} $50$ times and output the one that has the minimum objective value.
\vs
\noin
In algorithm above, $(\alpha^*, \lambda^*)$ is assumed to be selected. Now we address the issue how to choose this pair. Obviously, we can search among a finite grids over the region
$[0, 1)\times (0, \lambda_0]$.
\vs
\noin
\tb{Choice of the penalty/tuning parameters via cross-validation}.
\vs

We first pick a (relatively small) grid of values for $\lambda^*$, say from $0$ (excluded) to $\lambda_0$ with $\lambda_0/10$ as the step so that there are $10$ equal spaced grid points.
For the estimation of $\lambda_0$, one can see \cite{EHJT04} and  \cite{ACG13}, or \tb{(ii)} of Re-parametrization above.\vs
For each $\lambda^*$,  
we will
select an $\alpha$ value among $10$ equal spaced grid points over $[0,1)$ via five-fold cross-validation (CV).  In $k$-fold cross-validation, the data are split randomly in $k$ blocks (folds) of
approximately equal size. Each block is left out once to fit the model, and the left-out block is used as test data (see Section 7.10.1 of \cite{HTF17}). 
\vs
The CV is a popular method for estimating the prediction error and comparing different models (see \cite{ZH05} and \cite{HTF17}). The popular \textbf{R} package \emph{glmnet} can be used to select the parameters
as did in \cite{KHF18}, which automatically checks the model quality for a sequence of values for $\alpha$,
taking the mean squared error as an evaluation criterion.\vs
 We use a 5-fold CV via our own developed program to avoid the drawback of the glmnet which often leads to the error message ``from glmnet C++ code (error code 7777); All used predictors have zero variance" (this especially is true under the adversary contamination scenario). The latter leads to a problem for evaluating the performance of the procedure enetLTS of \cite{KHF18} when the contamination at $10\%$ level in next section. We have to drop enetLTS in the comparison in that situation.
We will ran ten times of our 5-fold CV, then the pair $(\lambda^*, \alpha)$ with   
the minimum averaged CV error will be the final chosen pair $(\lambda^*, \alpha^*)$.

\vs
The lars algorithm can be used to fit a linear model based on the $k-1$ blocks to obtain a $\widehat{\bs{\beta}}(\lambda^*, \alpha)$. Other algorithms, such as coordinate descent algorithms (including Fu's shooting algorithm (\cite{F98}) (see 2,11.1 of \cite{BVDG11}) can be employed to speed up the computation.
\vs
\noin

\section{Illustration examples and comparison \label{sec.7}}
\subsection{Simulation}
 All R code for simulation and examples as well as figures in this article (downloadable via https://github.com/left-github-4-codes/lst-enet) were run on a desktop Intel(R)Core(TM)
21 i7-2600 CPU @ 3.40 GHz.\vs
\noin
\tb{Five regularized regression procedures}
We like to compare the performance of our procedure lst-enet with leading regularized regression procedures including lasso, lars, enet, and enetLTS.  lasso will be computed via \textbf{R} package ``lars", it can be obtained via ``elasticnet". The latter package is  mainly for the enet whereas the former mainly focuses on lars.
Though lasso could be obtained via ``glmnet" but due to the contamination scenario, the glmnet often does not work. Unfortunately, enetLTS employing glmnet in its CV calculation, it can not hand the
model $y=\bs{w}'\bs{\beta}_0+{e}$ appeared in (\ref{model-population.eqn}) 
(an error message ``glmnet fails at standardization step''). We use an alternative model given below 
\vs

\noin
\tb{Simulation designs}
To copy with the situation above,
we simulate data from the true model:
$ y=\bs{X}\bs{\beta}_0 +\sigma e, ~~ e \sim N(0, 1),\label{model.22} $
where the true unknown parameter $\bs{\beta}_0$ is assumed to be a $p$-dimensional vector with the first $p_1:= \lceil6\%*p \rceil$ components are ones and the rest $p_2:=p-p_1$ components are zeros. $\sigma$ is set to be $0.5$ but could be changed to other values (leading to different signal-to-noise ratio).
\vs
\noin
\tb{Design I}: take sample of $\bs{X}$ from $N(\bs{0}, \sigma\bs{I}_{p\times p})$ and $e$ from $N(0, 1)$. \tb{Design II}: take sample from
 $\bs{X} 
 \sim N\big(\bs{0}, \bs{\Sigma}\big)$ with $\bs{\Sigma}(i,j)=\rho_1^{|i-j|}, 1\leq i, j \leq p_1$, $\bs{\Sigma}(i,j)=\rho_2^{|i-j|}, p_1< i, j \leq p$, $\rho_1=0.95$, $\rho_2=0.05$, all other entries of $\bs{\Sigma}$ are zeros and $e\sim N(0,1)$. We  take $n\in \{50, 100\}$ samples from the $\bs{X}$ and $e$ above and calculate the response $y_i=\bs{X}_i\bs{\beta}_0+\sigma e_i, i\in \{1, 2, \cdots, n\}$.
 \vs
\noin
\tb{Contamination levels and schemes}
Let $\ep$ be the contamination level, when $\ep=0$ there is no contamination, an ideal situation (and not realistic). Consider the scenario  $\ep \in \{0, 0.05, 0.1, 0.2\}$ (i.e., $0\%$, $5\%$, $10\%$, $20\%$ contamination). Let $m=\lfloor \ep*n\rfloor$, sample $m$ indices from $\{1,\cdots, n\}$. \vs  Contamination \tb{Scheme I}: add $20$ to the corresponding $m$ components of $(e_1, \cdots, e_n)$, compute $y_i=\bs{X}_i\bs{\beta}_0+\sigma e_i, i\in \{1, 2, \cdots, n\}$, and add $20$ (component-wise) to the
corresponding $m$ rows of $(\bs{X}_1, \cdots, \bs{X_n})'$.  \tb{Scheme II}: add $20$ to the corresponding $m$ components of $(e_1, \cdots, e_n)$, compute $y_i=\bs{X}_i\bs{\beta}_0+\sigma e_i, i\in \{1, 2, \cdots, n\}$. Replace the corresponding $m$ rows of $(\bs{X}_1, \cdots, \bs{X_n})'$ by a p-vector with its first component being $10^4$ and the rest are zeros, do the same for the corresponding $m$ components of $(y_1, \cdots, y_n)$ but with a scalar $10^{10}$. 
\vs
\noin
\tb{Four performance criteria}
The first  measure is the estimation error, or L2-error/L2-loss between the true parameter $\bs{\beta}_0$  and the estimator $\widehat{\bs{\beta}}_P$ via procedure $P$ and is defined as:
\be
 \mbox{L2-error}(\bs{\beta}_0, \widehat{\bs{\beta}}_P):=\|\bs{\beta}_0-\widehat{\bs{\beta}}_P\|^2_2,
\ee
where $\|\bs{a}-\bs{b}\|_2$ is  the $\ell_2$-norm between the two p-dimensional vectors.
\vs
On the other hand, one has to take the performance measure into the context of the sparsity model consideration. In the following we introduce the \emph{true sparsity discovery rate} (TSDR) and the \emph{false sparsity discovery rate} (FSDR). For notation simplicity, we denote the unknown parameter by $\bs{\beta}^{0}$ (assume it has at least one zero coordinate), an estimator by $\widehat{\bs{\beta}}^P$.
\be
\mbox{TSDR}(\bs{\beta}^0, \widehat{\bs{\beta}}^P):=\frac{\sum_{i=1}^p\mathds{1}(\beta^0_i=0, \hat{\beta}^P_i=0)}{\sum_{i=1}^p\mathds{1}(\beta^0_i=0)},
\ee
namely, the fraction of correctly detecting/discovering the zero coordinates of the true parameter $\bs{\beta}^0$. The higher the TSDR, the better the $\widehat{\bs{\beta}}^P$.
\be
\mbox{FSDR}(\bs{\beta}^0, \widehat{\bs{\beta}}^P):=\frac{\sum_{i=1}^p\mathds{1}(\beta^0_i\not =0, \hat{\beta}^P_i=0)}{\sum_{i=1}^p\mathds{1}(\beta^0_i\not =0)},
\ee
namely, the fraction of falsely detecting/discovering as zero coordinate for the true parameter $\bs{\beta}^0$. The lower the FSDR, the better the $\widehat{\bs{\beta}}^P$.
\vs

The fourth performance measure is a popular one, it is (square-)root of mean squared (prediction) error (RMSE) on testing data. That is, for a given data set, one first partitions data into training and testing two parts (we take the ratio 7:3 for partition). Then fit the model and get estimator based on the training data and using the testing data to get the RMSE. Testing data sets are often assumed to be clean (have no contamination or outliers) in the literature. This, however, is not realistic in practice. \vs

\vspace*{-0mm}

Let $\bs{X}_{test}$, $y_{test}$ be the testing data and $\widehat{\bs{\beta}}_P$ be the estimator obtained from the training data. Then
\be
\mbox{RMSE}(\widehat{\bs{\beta}}_P):=\Big(\mbox{mean}\big((y_{test}-\bs{X}_{test} \widehat{\bs{\beta}}_P)^2\big)\Big)^{1/2}.
\ee
\bec
\begin{figure}[h!]
\vspace*{-7mm}
\includegraphics
[width=\textwidth]
{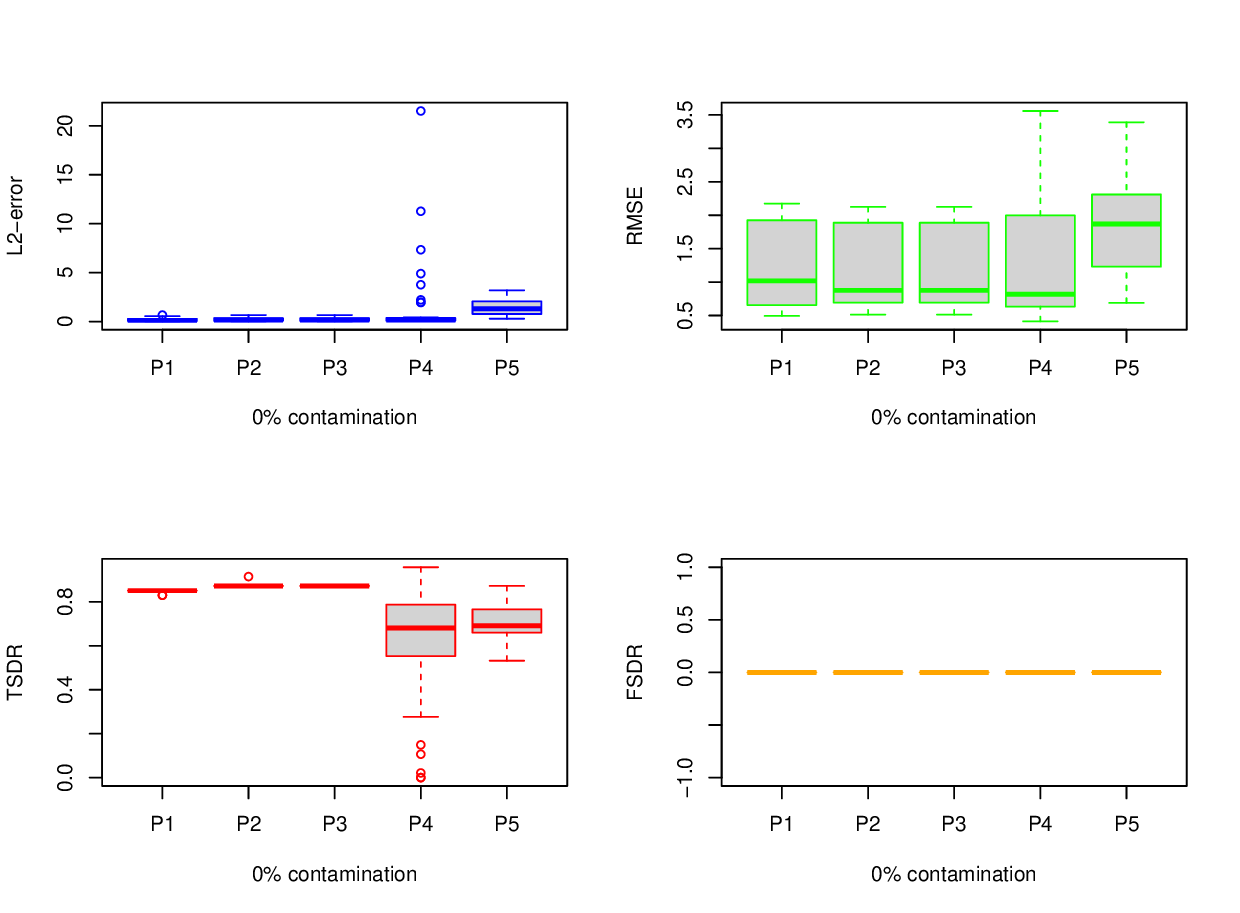}
 \caption{\scriptsize  Boxplots for five procedures (P1 stands for lst-enet, P2 for lasso, P3 for lars, P4 for enetLTS, P5 for enet) and 50 samples each with $n=100$ and $p=50$ that are generated from design I with $0\%$ contamination rate.}
 \label{fig-box1}
\vspace*{-0mm}
\end{figure}
\enc
\vspace*{-8mm}
\noin
The four performance measures above were discussed in the literature before, all are hoped to be small except the TSDR which is hoped to be as high as possible. All (but RMSE) depend on the unknown parameter $\bs{\beta}^0$. 
\vs
\noin
\tb{Example 7.1}
We first consider $\ep=0$. For simplicity,  data are generated according to design \tb{I} and set $n=100$, $p=50$ (low dimension case) or $n=50$, $p=300$ (high dimension and sparsity case). We generated 50 samples for $\bs{X}$ and $e$ and obtained corresponding responses $y$. The simulation results are displayed in Figure \ref{fig-box1}. For description simplicity, we use hereafter P1 for lst-enet, P2 for lasso, P3 for lars, P4 for enetLTS, P5 for enet in the Figures.

\vs
Inspecting Figure \ref{fig-box1} reveals that (i) with respect to (w.r.t.) FSDR, all four perform equally well with $0\%$ mis-discovery rate; (ii) w.r.t. TSDR, lst-enet, Lars, and lasso perform stably and at a highest rate while enet with a  less stable lower rate but enetLTS performs most unstable with the lowest median rate; (iii)w.r.t. RMSE, lasso and lars are the best followed by lst-enet, enetLST has the median RMSE that is also among the best but with the widest spread of RMSE  while
enet has the largest (and wider spread of) RMSE; (iv) w.r.t. L2-error, lst-enet, lars, and lasso are among the best while enetLTS has the worst performance followed by enet. Overall, lst-enet, lars, and lasso are among the best whereas enetLTS performs worst overall followed by enet.
\vs
\noin
\bec
\begin{figure}[h!]
\vspace*{-8mm}
\includegraphics
[width=\textwidth]
{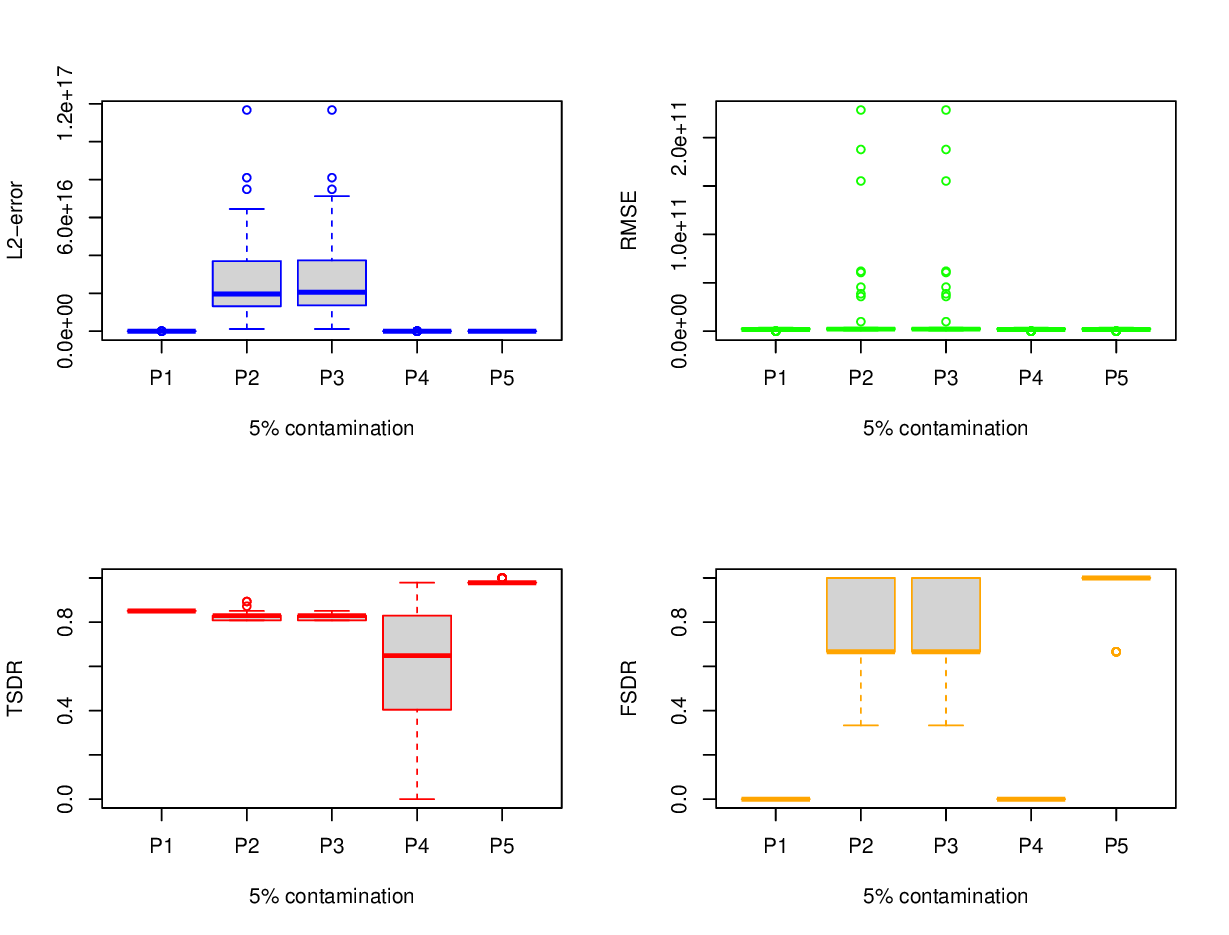}
 \caption{\scriptsize  Boxplots for five procedures 
 and 50 samples each with $n=100$ and $p=50$ that are generated from design I with $5\%$ contamination rate.}
 \label{fig-box33}
\end{figure}
\enc
\vspace*{-8mm}

\noin
\tb{Example 7.2} Perfect normal data are not realistic in practice.
We  now consider $\ep=0.05$ (i.e. $5\%$ contamination), all others are the same as Example 7.1 except the contamination scheme II will be adopted (in 7.1 contamination scheme does not matter).
We first consider $n=100$, $p=50$ (low dimension case) and for simplicity generate data according design \tb{I}. Performance of five procedures in 50 samples is displayed in Figure \ref{fig-box33}.
\bec
\begin{figure}[h!]
\vspace*{-3mm}
\includegraphics
[width=\textwidth]
{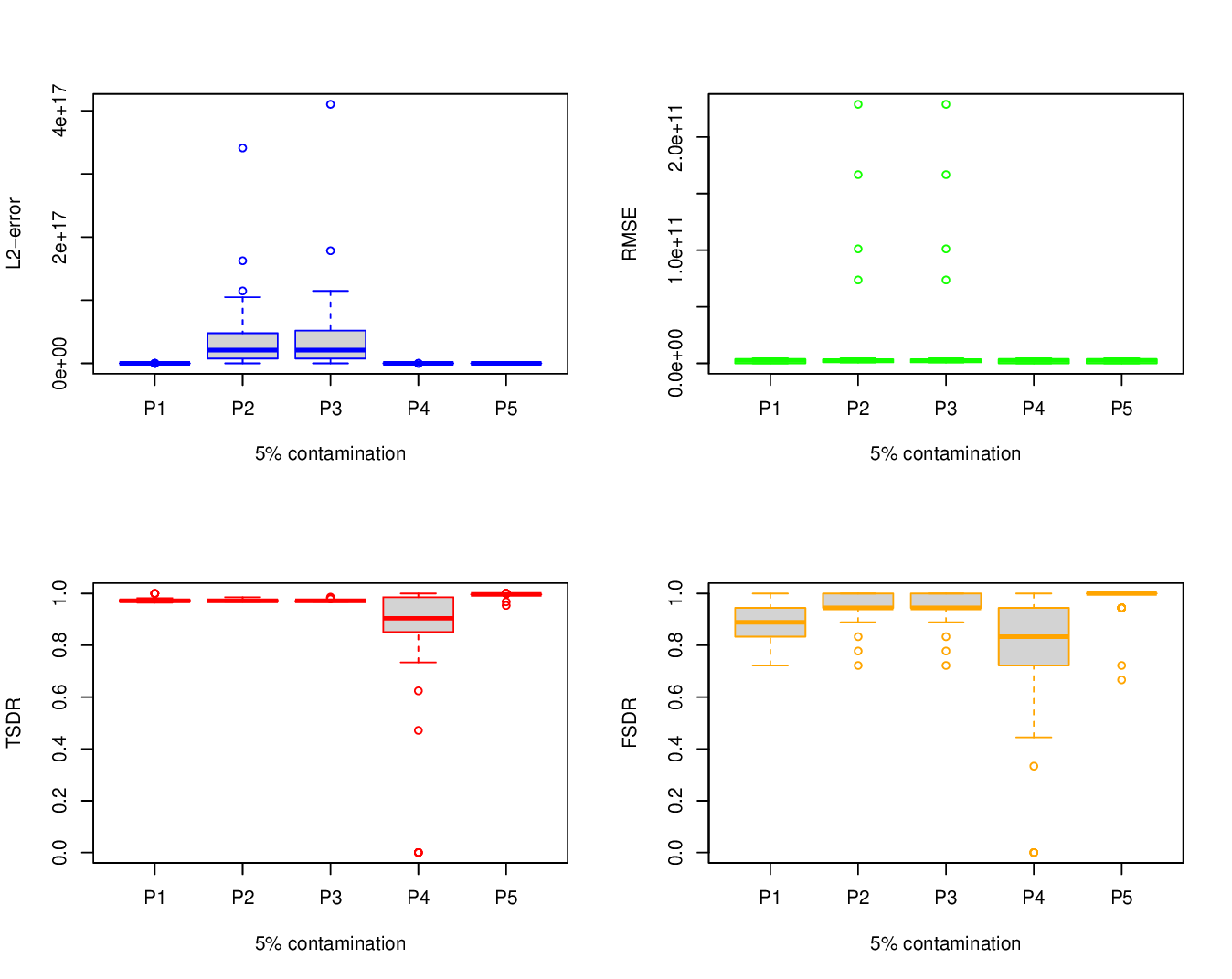}
 \caption{\scriptsize  Boxplots for five procedures 
 and 50 samples each with $n=50$ and $p=300$ that are generated from design I with $5\%$ contamination rate.}
 \label{fig-box4}
\end{figure}
\enc
\vspace*{-8mm}

Inspecting the Figure reveals that (i)  w.r.t. L2-error, lst-enet, enetLTS and enet are the best performers while lasso and lars are equally dissatisfactory;
(ii)  w.r.t. RMSE, the situation is the same as in the L2-error case; (iii)  w.r.t. TSDR, enet is the best performer (this perhaps is false best since it might assign zero to all components of the estimator $\widehat{\bs{\beta}}$ that could lead to $100\%$ of its FSDR) while enetLTS is the worst; (iv)  w.r.t. FSDR, lst-enet, enetLTS are the best performers followed by lasso and lars while enet is the loser. Overall, lst-enet is the only winner.
\vs
The simulation study above with $5\%$ contamination is repeated but $n=50$ and $p=300$ (high dimensional case) and simulation design II is adopted. Results are displayed in Figure  \ref{fig-box4}.
\vs
Reviewing the Figure discovered that (i)  w.r.t. L2-error,  lst-enet, enetLTS, and enet are the best performers while lasso and lars are disappointing; (ii)  w.r.t.  RMSE;
the situation is almost the same as in L2-error case;
(iii)  w.r.t. TSDR, lst-enet, lasso, lars  and enet are the best performers while enetLTS is the loser;
(iv)  w.r.t. FSDR,  enet is the worst performer (since its FSDR is almost $100\%$), enetLTS has the lowest median value while it has the widest spread. lst-enet is the second best performer, lasso and lars are disappointed. Overall, lst-enet is the only winner.
\vs
\bec
\begin{figure}[h!]
\vspace*{-5mm}
\includegraphics
[width=\textwidth]
{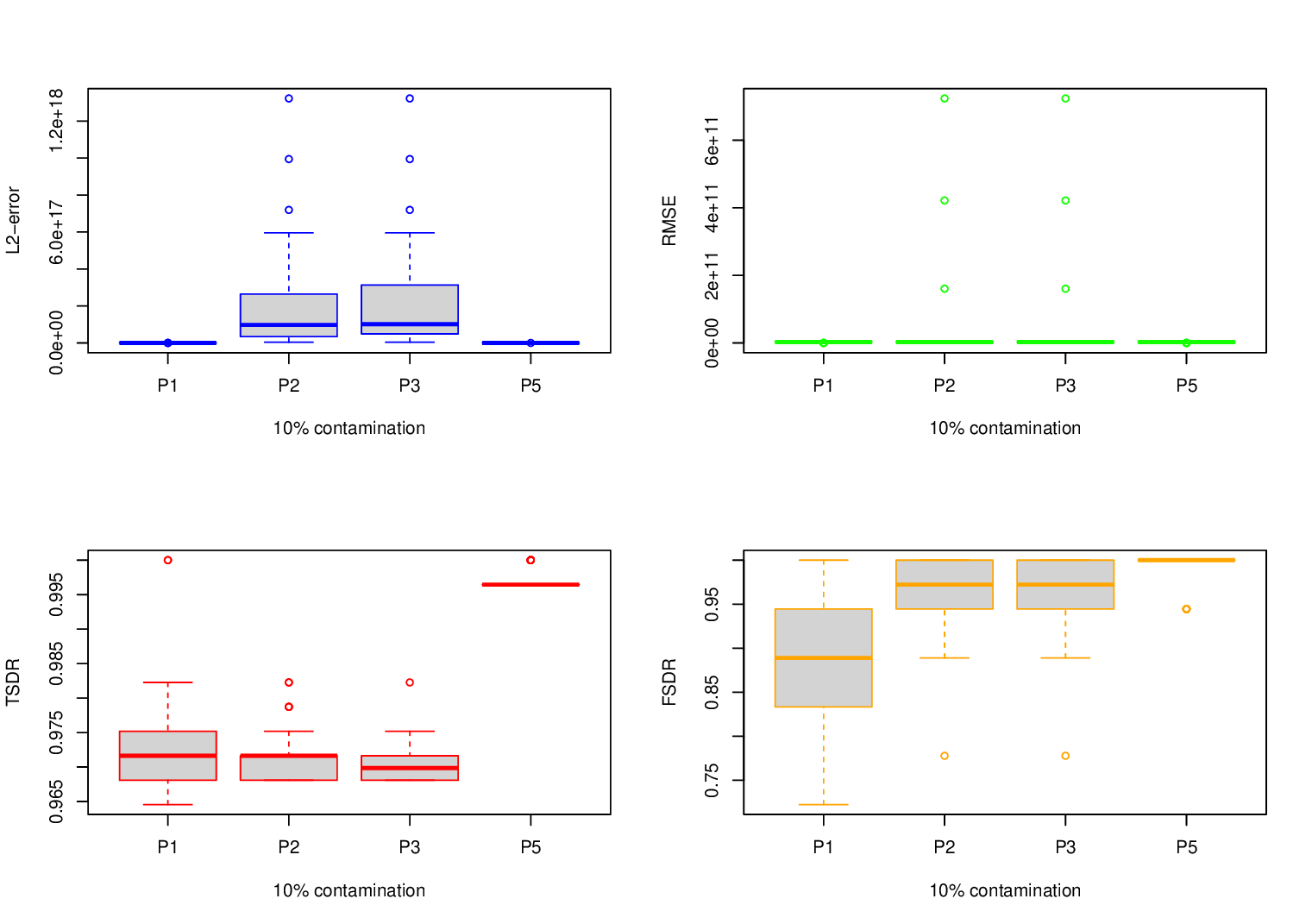}
 \caption{\scriptsize  Boxplots for four procedures
 and 50 samples each with $n=50$ and $p=300$ that are generated from design I with $10\%$ contamination rate.}
 \label{fig-box5}
\end{figure}
\enc
\vspace*{-8mm}
\noin
\tb{Example 7.3}
In practice, $10\%$ (or even $20\%$) contamination is not rare. Next we consider the case $\ep=0.1$ (i.e., $10\%$ contamination), contamination scheme II will be adopted. Samples of 50 with $n=50, p=300$ are generated with simulation design I. Due to the higher level contamination and the usage of \textbf{R} package glmnet in its background CV calculation,  enetLTS fails to go through the computation we have to drop it in our comparison.
Simulation results are displayed in Figure  \ref{fig-box5}.\vs

\begin{figure}[h!]
\vspace*{-2mm}
\includegraphics
[width=\textwidth]
{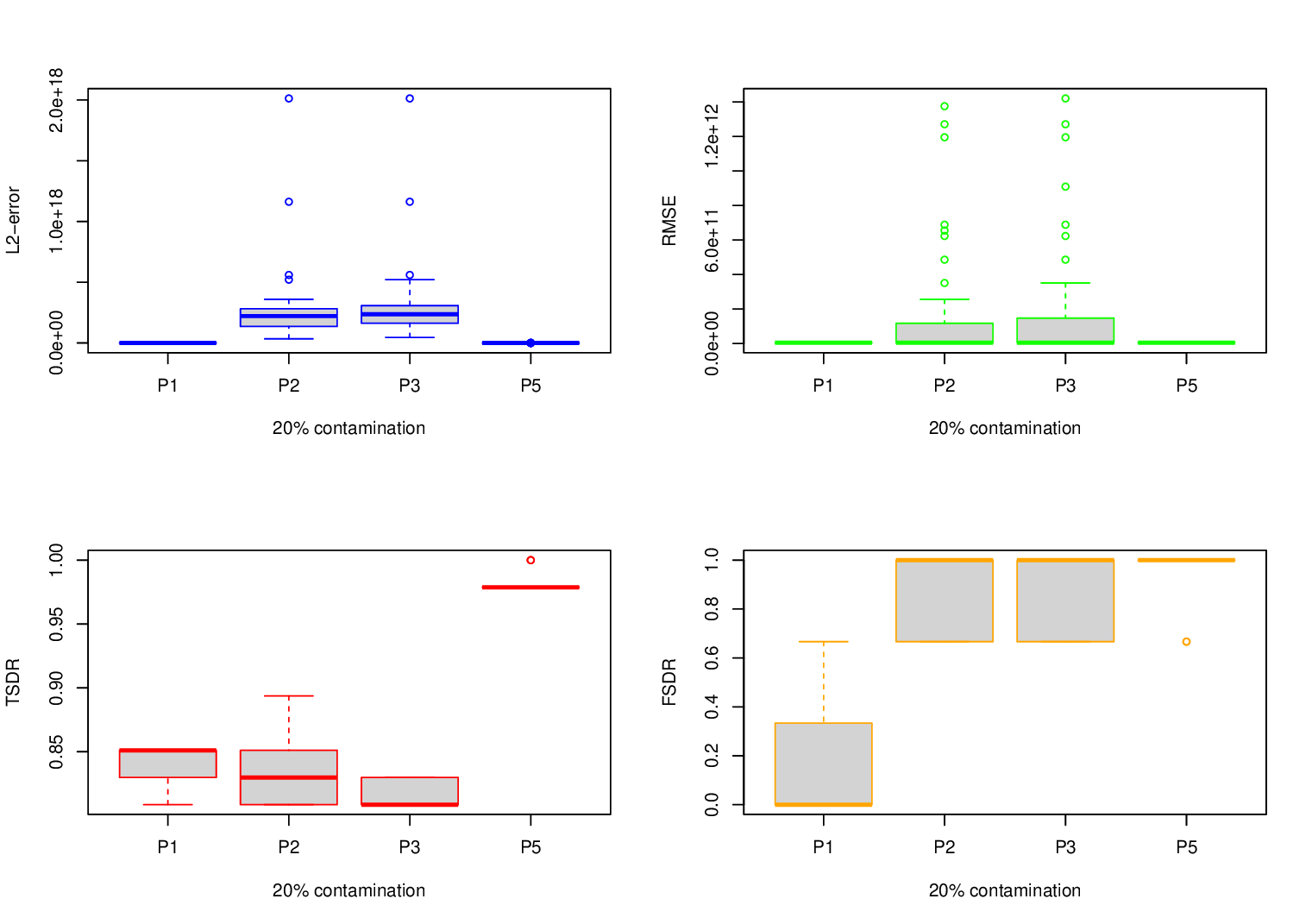}
 \caption{\scriptsize  Boxplots for four procedures
 and 50 samples each with $n=100$ and $p=50$ that are generated from design I with $20\%$ contamination rate.}
 \label{fig-box6}
\end{figure}

Inspecting the Figure reveals that (i)  w.r.t. L2-error, lst-enet and enet are the best while lasso and lars are inferior; (ii)  w.r.t. RMSE, the situation is the same as in L2-error case; (iii)  w.r.t. TSDR, enet is the worst performer (it assigns zeros to almost all components of $\widehat{\bs{\beta}}$ that will lead to $100\%$ of its FSDR), lst-enet and lasso are the best performers followed by lars; (iv)  w.r.t. FSDR, lst-enet is the best performer, enet is the worst one while lasso and lars perform dissatisfactory. Overall, lst-enet is the winner.
\vs
The advantage of lst-enet is even better demonstrated in Figure \ref{fig-box6} when $n=100$ and $p=50$ and $\ep=0.2$ (i.e., $20\%$ contamination).
\bec
\enc

\vspace*{-8mm}

\subsection{A read data example}
\noin
\tb{Example 7.4}
To analyze a realistic dataset with very large number of variables,
we consider the well-known cancer data  from the National Cancer
Institute (NCI60); see \cite{Tetal09} 
 for more detail about this dataset. 
 A total of 59 of
the human cancer cell-lines (n= 59) were assayed for gene
expression and protein expression. The data set,  downloadable from the CellMiner program package, NCI (http://discover.nci.nih.gov/cellminer/) and available from the \textsf{R} package robustHD, has been repeatedly studied in the literature, see e.g., \cite{LLLP11}.

\bec
\begin{figure}[h!]
\vspace*{-5mm}
\includegraphics[width=11cm, height=8cm]
{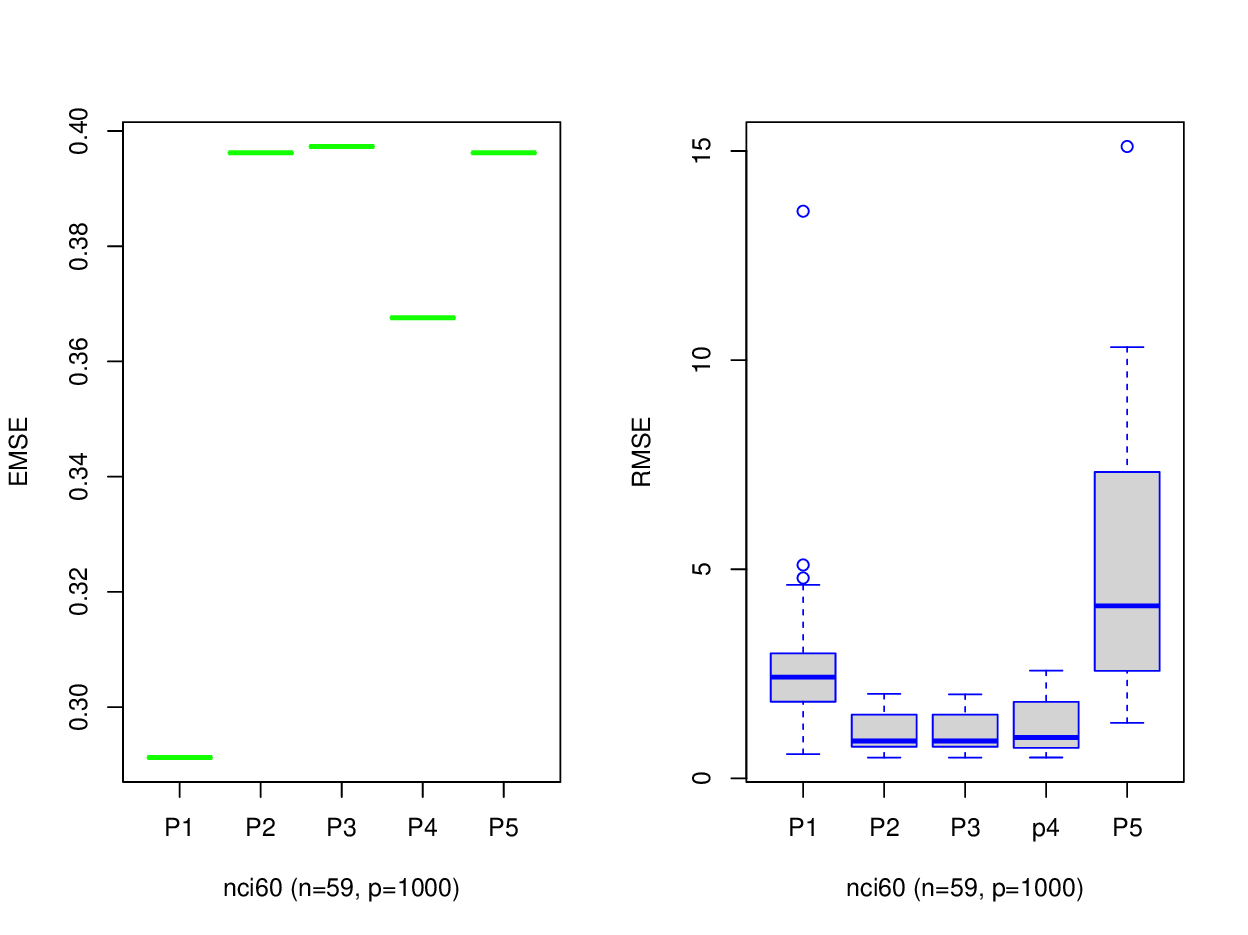}
 \caption{\scriptsize  Boxplots for five procedures 
  for real data set NCI60 with 50 times partition of a selected sub-dataset $p=1000$ and $n=59$.}
 \label{fig-box7}
\vspace*{-3mm}
\end{figure}
\enc
\vspace*{-10mm}

 We process the data set by following the approach in the literature and treat the gene expression microarray data as the predictors $\bs{X}_{raw}$ (a $59$ by $22283$ matrix) and the protein expression data as responses variables $\bs{Y}_{raw}$ ( a $59$ by $162$ matrix). Similar to \cite{LLLP11} or \cite{ACG13}, we order the protein expression variables according to their scale (employing MAD as a scale estimator instead of the standard deviation) and select the one with median 
  MAD, serving as our dependent variable. It is 75the column 
   of the protein expression data matrix. Denote it by $\bs{Y}$. 
 \vs Next, we
selected out genes using their correlations with $\bs{Y}$. Here we adopt the robust correlation measure in \cite{KVZ07}. We obtain $22283$ ordered (decreasing) correlations and select top
$k1=100$ corresponding columns of $\bs{X}_{raw}$ and combined with the bottom $1000-k1$ columns as our final $\bs{X}$, reducing the  number of
genes from 2,2283 to $p=1,000$. The number $p$ could easily be changed by adjusting $k1$. \vs

We partition (by rows) $\bs{X}_{59 \times 1000}$  and $\bs{Y}$ into $x.train$, $y.train$ and $x.test$, $y.test$ according the rate $7:3$. That is $41$ rows of $\bs{X}$ and $\bs{Y}$ for the training data sets, the rest $18$ rows as testing data sets. We do this step 50 times and each time we calculate the RMSE (the only measure that still valid without the given $\bs{\beta}_0$) for the five procedures. The results are displayed in Figure \ref{fig-box7}, where the fifth performance measure is introduced, that is, the empirical mean squared error defined as:
\be
\mbox{EMSE}(\widehat{\bs{\beta}}_P):=\frac{1}{R}\sum_{i=1}^R\big(\widehat{\bs{\beta}}_P-\overline{\widehat{\bs{\beta}}}_P\big)^2, \label{emse.eqn}
\ee
where $R$ is the replication number, namely, $\frac{R}{R-1}\mbox{EMSE}$ is the sample variance of $\widehat{\bs{\beta}}_P$. $\overline{\bs{x}}$ stands for the sample mean of $\bs{x}_i$.
\vs
Inspecting the Figure reveals that (i) lasso, lars, enetLTS (enetLTS has the wider spread) have the smallest RMSE but their sample variances (EMSE) are among the largest; (ii)  RMSE of the lst-enet is the second smallest but it is the most stable estimator with the distinguished smallest sample variance (EMSE) which means that with different training and testing data sets obtained by random partitioning, lst-enet produces very closed solutions; (iii) enetLTS has the lowest median RMSE but its sample variance is the remarkably large; (iv)  enet has the categorical largest RMSR while its sample variance is also the largest. Overall, lst-enet  is recommended with the rivals enetLTS, lasso and lars.
\section{Concluding discussions \label{sec.8}}

Most of leading penalized regression estimators for high-dimensional sparse data can breakdown by a single outlier (or contaminating point). The newly proposed lst-enet estimator not only processes a high breakdown robustness but also performs well in simulation studies and a read data example, serving as a robust alternative to regularized regression estimators.\vs
\noin
\tb{Robust measure} Finite sample breakdown point has been served as a prevailing quantitative robustness measure in finite sample practice, the main advantage/beauty is its non-randomness and probability-free nature that is exactly why it was enthusiastically welcomed and quick became adopted in a broad spectrum of disciplines after its introduction in 1983.
\vs
 Critics (e.g., \cite{SWS22}), however, would like to have a more complicated version, a version that includes randomness and Orlicz norm. They argued that worst case performance might not be a good robustness measure. On the other hand, it is common practice to use the worst case performance as in the complexity of an algorithm or the safety of a passenger cars case. \vs
\noin
\tb{Future possible work}~~
 (a) Further performance measure could be pursued including  \tb{(i)} whether $\hat{\bs{\beta}}$ performs well on future samples (i.e., whether
$ E( Y-\bs{w}'\hat{\bs{\beta}})^2$ is small);
\tb{(ii)} whether $\hat{\bs{\beta}}$ closely approximates the ``true" parameter $\bs{\beta}_0$ (i.e., whether
$\|\hat{\bs{\beta}}-\bs{\beta}_0\|$ is small with high probability); or
\tb{(iii)}
{whether} $\hat{\bs{\beta}}$ correctly identifies the relevant coordinates of the ``true,"
sparse parameter $\bs{\beta}_0$ (i.e., whether ($\bs{\beta}_{0j}= 0)	\Leftrightarrow( \hat{\bs{\beta}}_j = 0)$ with high
probability). 
(b) Extension of current regression work  to a more general setting to cover  discriminant analysis, logistic regression, and other topics. \vs
  \vs
\begin{center}
{\textbf{\large Acknowledgments}}
\end{center}
\vs
The author thanks 
Prof.s Haolei Weng, Yiyuan She  and Wei Shao for
 insightful comments and stimulating discussions which significantly improved the manuscript.
\vs
\vs
\noindent
{\Large \tb{Appendix: proofs of main results}}
\vs

\noindent
\tb{Proof of Theorem 2.1}
\vs
\noin
\tb{Proof}: Clearly, it suffices to show that RBP$(\widehat{\bs{\beta}}^*(\lambda_1, \lambda_2,\gamma, \bs{Z}^{(n)}), \bs{Z}^{(n)})\leq 1/n$. Equivalently, to show that one point can break down the estimator.
Assume, otherwise, one point is not enough to break down the estimator. That is, there exists an $M$ such that
\be\sup_{\bs{Z}^{(n)}_1} \| \widehat{\bs{\beta}}^*(\lambda_1, \lambda_2,\gamma, \bs{Z}^{(n)}_1)\|_2 <M<\infty, \label{0-inequality.eqn}\ee
where $\bs{Z}^{(n)}_1$ stands for any contaminated data set by replacing one point in the original data set $\bs{Z}^{(n)}$ with an arbitrary point in $\R^p$.
We seek a contradiction now.
\vs
 Replace $\bs{Z}_1=(\bs{x}'_1, y_1)'$ in $\bs{Z}^{(n)}=\{\bs{Z}_1, \cdots, \bs{Z}_n\}$ by $\bs{Z}_1^*=((\delta, 0,\cdots, 0), \kappa \delta )'$.
 Denote the contaminated data set by $\bs{Z}^{(n)}_1$ and the estimator based on it as $\widehat{\bs{\beta}}^*:=\widehat{\bs{\beta}}^*(\lambda_1, \lambda_2,\gamma, \bs{Z}^{(n)}_1)$. \vs
 Let $M_y=\max_{i}|y_i|, M_x=\max_{i}|\bs{x}_{i1}|$.
Let $\bs{\beta}_{\kappa}=(0, \kappa, 0, \cdots, 0)'\in \R^p$ and set $\kappa=(\sqrt{p}+1)M+1$.
Select a large $\delta$ such that   
 $\mc{L}(\delta)/n\geq \mc{L}(M_y+M_x\kappa)+g(\bs{\beta}_{\kappa}, \lambda_1, \lambda_2, \gamma)+1$,
 This is possible since $\mc{L}(x)\to \infty$ when $|x|\to \infty$ and $\mc{L}(x)$ is non-decreasing over $(0, \infty)$.
 Then 
\begin{align}
O(\widehat{\bs{\beta}}^*)\leq O(\bs{\beta}_{\kappa})&=\frac{1}{n}\sum_{i=1}^n \mc{L} (r_i)+g(\bs{\beta}_{\kappa}, \lambda_1, \lambda_2, \gamma) \nonumber\\[1ex]
 &=\frac{1}{n}\sum_{i=2}^n \mc{L} (r_i)+g(\bs{\beta}_{\kappa}, \lambda_1, \lambda_2, \gamma)~~ ~\mbox{(since $\mc{L}(r_1)=\mc{L}(0)=0$)} \nonumber\\[2ex]
 &\leq \frac{n-1}{n}\mc{L}(M_y+M_x\kappa)+g(\bs{\beta}_{\kappa}, \lambda_1, \lambda_2, \gamma) \nonumber\\[1ex] %
 &\leq \frac{1}{n}\mc{L}(\delta)-1.
 \label{1st-inequlity.eqn}
\end{align}
 On the other hand, for any $\bs{\beta} \in \R^p$ such that $(\sqrt{p}+1)\|\bs{\beta}\|_2\leq\kappa-1$, one has
 \begin{align}
 O(\bs{\beta})&\geq \frac{1}{n}\mc{L}(y_1-\bs{w}_1\bs{\beta})
 = \frac{1}{n}\mc{L}(\kappa \delta-(\beta_1+\delta\beta_2))\nonumber\\[1ex]
 &\geq \frac{1}{n}\mc{L}(\kappa \delta-(\delta|\beta_1|+\delta|\beta_2|))\nonumber\\
&\geq \frac{1}{n}\mc{L} (\delta(\kappa- (\sqrt{p}+1)\|\bs{\beta}\|_2))\geq \frac{1}{n}\mc{L} (\delta), \label{2nd-inequality-1.eqn}
\end{align}
where the facts: (i) $|\beta_i|\leq \|\bs{\beta}\|_i (i=1,2)$ and (ii) $\|\bs{\beta}\|_1\leq \sqrt{p}\|\bs{\beta}\|_2$ are utilized.

Combining 
(\ref{1st-inequlity.eqn}) and (\ref{2nd-inequality-1.eqn}), leads to the conclusion that
$$\|\widehat{\bs{\beta}}^*(\lambda_1, \lambda_2,\gamma, \bs{Z}^{(n)}_1)\|_2 > \frac{\kappa-1}{\sqrt{p}+1}= M,   $$
which contradicts (\ref{0-inequality.eqn}). \hfill \pend

\vs
\noindent
\tb{Proof of Theorem 4.1}
\vs
\noindent
\tb{Proof}:
 $\lambda_1+\lambda_2=0$ case has been treated in \cite{ZZ22}, we treat $\lambda_1+\lambda_2>0$ case here.\vs
\tb(i) Denote the objective function on the RHS of (\ref{lst-en.eqn}) as
\be O_n(\bs{\beta},\lambda_1, \lambda_2, \gamma):=O(\bs{\beta}, \lambda_1, \lambda_2, \alpha,  \gamma, \bs{Z}^{(n)})=\frac{1}{n}\sum_{i=1}^{n}r_i^2w_i+\lambda_1 \sum_{j=1}^p|\beta_j|^{\gamma}+\lambda_2\sum_{i=1}^p\beta^2_i.\label{objective.eqn}
\ee
Denote the three terms on the RHS above as $g(\alpha, \bs{\beta}, \bs{Z}^{(n)})$, $g_1(\lambda_1, \gamma, \bs{\beta})$, and $g(\lambda_2,\bs{\beta})$, respectively. It is readily seen that
the RHS of (\ref{lst-en.eqn}) is equivalent to minimizing $G(\bs{\beta}, \alpha, \lambda_1, \gamma):=g(\alpha, \bs{\beta}, \bs{Z}^{(n)})+g_1(\lambda_1, \gamma, \bs{\beta})$ subject to $\sum_{i=1}^p\beta^2_i \leq t_2$, $t_2\geq 0$ 
\vs
By Lemma 2.2 of \cite{ZZ22}, $g(\alpha, \bs{\beta}, \bs{Z}^{(n)})$ is continuous in $\bs{\beta}$ (this is not as obvious as one believed) while the continuity of $g_1(\lambda_1, \gamma,  \bs{\beta})$
in $\bs{\beta}$ is obvious. Therefore we have a continuous function of $\bs{\beta}$, $G(\bs{\beta}, \alpha, \lambda_1, \gamma)$, 
which obviously has minimum value over the compact set $\|\bs{\beta}\|_2\leq t_2$. 
\vs
{(ii)} Follows the approach originated in \cite{ZZ22}, we partition the parameter space $\R^p$ of $\bs{\beta}$ into disjoint open pieces $R_{\bs{\beta}^k}$, $1\leq k\leq L\leq {n \choose \lfloor (n+1)/2\rfloor}$ and $\cup_{1\leq k\leq L}\overline{R}_{\bs{\beta}^k}=\R^p$, where $\overline {A}$ stands for the closure of the set $A$, and
\be
R_{\bs{\beta}^k}=\{\bs{\beta}\in \R^p: I(\bs{\beta})=I(\bs{\beta}^k), D_{i_1}(\bs{\beta})< D_{i_2}(\bs{\beta})\cdots< D_{i_K}(\bs{\beta})\}, \label{region.eqn}
\ee
where $D_i:=D(r_i, \bs{\beta})={|r_i-m(\bs{Z}^{(n)},\bs{\beta})|}\big/{\sigma(\bs{Z}^{(n)},\bs{\beta})}$ for a given $\mb{Z}^{(n)}$ and $\bs{\beta}$,
$i_1,\cdots, i_{K}$ in $I(\bs{\beta})$ 
and $K=|I(\bs{\beta})|$ with $w_i$ defined in (\ref{lst-en.eqn})
\be
I(\bs{\beta})=\Big\{ i:  w_i=1 
\Big\}. \label{I-beta.eqn}
\ee
For any $\bs{\beta} \in \R^p$, either there is $R_{\bs{\eta}}$ and
$\bs{\beta} \in R_{\bs{\eta}}$ or there is $R_{\bs{\xi}}$, such that $\bs{\beta} \not\in R_{\bs{\eta}}\cup R_{\bs{\xi}}$ and $\bs{\beta}\in \overline{R}_{\bs{\eta}}\cap \overline{R}_{\bs{\xi}}$. 
Now we claim that  $ \widehat{\bs{\beta}}:=\widehat{\bs{\beta}}^n_{lts-enet}(\alpha, \lambda_1, \lambda_2, \gamma)\in R_{\bs{\beta}^{k_0}}$ for some $1\leq k_0\leq L$.
\vs
Otherwise, assume that
$\widehat{\bs{\beta}} \in \overline{R}_{\bs{\beta}^{k_0}}$. By Lemma 2.2 of \cite{ZZ22}, $g(\alpha, \bs{\beta}, \bs{Z}^{(n)})$ (denoted by $Q^{n}(\bs{\beta})$ there) is 
convex over 
 $R_{\bs{\beta}^{k_0}}$. Therefore, $O_n(\bs{\beta},\lambda_1, \lambda_2, \gamma)$ is strictly convex in $\bs{\beta}$ over  $R_{\bs{\beta}^{k_0}}$.  Assume that $\bs{\beta}^*$ is the global minimum of  $O_n(\bs{\beta},\lambda_1, \lambda_2, \gamma)$ over  $R_{\bs{\beta}^{k_0}}$. Then it is obviously that $O_n(\widehat{\bs{\beta}},\lambda_1, \lambda_2, \gamma)\leq O_n(\bs{\beta}^*,\lambda_1, \lambda_2, \gamma)$.
 But this is impossible in light of Lemma 4.1.
The strict convexity of $O_n(\bs{\beta},\lambda_1, \lambda_2, \gamma)$ over $R_{\bs{\beta}^{k_0}}$ guarantees the uniqueness.
 \hfill \pend
 \vs
 \noin
 \tb{Proof of Lemma 4.1}
 \vs
 \noin
\tb{Proof:}  Let $B(\bs{x}^*, r)$ be a small ball centered at $\bs{x}^*$ with a small radius $r$ and $  B(\bs{x}^*, r)\subset S$. Let $B^c:= S 
 -B(\bs{x}^*, r)$ and $\alpha^*=\inf_{\bs{x} \in B^c}f(\bs{x})$.
 Then, $\alpha^*>f(\bs{x}^*) $ (in light of strict convexity) 
 Since $y\in \overline{S}$, then there is a sequence $\{\bs{x}_j\} \in B^c$ such that $\bs{x}_j \to \bs{y}$ and $f(\bs{x}_j)\to f(\bs{y})$ as $j\to \infty$. Hence $f(\bs{y})=\lim_{j\to \infty}f(\bs{x}_j)\geq\alpha^*> f(\bs{x}^*)$. \hfill \pend
\vs

\noindent
\tb{Proof of Theorem 4.2}
\vs
\noindent
\tb{Proof}: We complete it in two steps.\vs
\noin
\tb{(i)} \emph{$m\leq n-k$ contaminating points  are not enough to break down the estimator}.
Let $M_y=\max_{i}|y_i|$, denote the minimizer of the Q in (\ref{objective-45.eqn}) for the contaminated sample as $\widehat{\bs{\beta}}$, Then,
 it is obviously that
$$
Q(\widehat{\bs{\beta}})\leq Q(\bs{0}, \lambda_1, \lambda_2, \gamma, Z^{(n)}_m)=\frac{1}{n}\sum_{i=1}^n r_i^2 w_i \leq \frac{k}{n} M^2_y,
$$
where the last inequality deserves further explanations. Note that there are at least $k$ un-contaminated (original) points.\vs Therefore, in the case that $w_i=\mathds{1}(r_i^2\leq r^2_{h:n})$,
the RHS of the above display wants to keep the sum of smallest $k=h$ squared residuals ($y^2_i$), this sum is certainly no greater than that of $k=h$ squared residuals from the $k$ original points.\vs
Likewise, in the case of $w_i=\mathds{1}(D(r_i, R^{(n)}\leq \alpha)$, the RHS wants to keep the sum of squared residuals ($y^2_i$) from $k=|I(\bs{0})|$ points that have the smallest outlyingness no greater than $\alpha$,
which is certainly no greater than the sum  of $k$ squared residuals from the $k$ original points.\vs
Assume, w.l.o.g. that $\lambda_1>0$ ($\lambda_2>0$ is even easier). Consider any $\bs{\beta}$ with $\|\bs{\beta}\|_2\geq M:=({(k+1)M^2_y}\big/{n\lambda_1})^{1/\gamma}$, then
\[
Q(\bs{\beta}, \lambda_1, \lambda_2, \gamma, Z^{(n)}_m)> \lambda_1\sum_{i=1}^p|\beta_i|^{\gamma}= \lambda_1\|\bs{\beta}\|^{\gamma}_{\gamma}\geq \lambda_1\|\bs{\beta}\|^{\gamma}_2\geq \frac{k+1}{n} M^2_y,
\]
where the fact that $\|x\|_q\leq \|x\|_p$ when $1\leq p\leq q <\infty$  is invoked.\vs
The  two displays above imply that
$$
\|\widehat{\bs{\beta}}(\lambda_1,\lambda_2,\gamma,\bs{Z}^{(n)}_m)\|_2< M.
$$
\vs
\noin
\tb{(ii)} \emph{$m= n-k+1$ contaminating points  are enough to break down the estimator}.\vs  The structure and basic idea of this part is an analogue to that of proof of Theorem 2.1.
Assume, otherwise, $m$ points are not enough to break down the estimator. That is, there exists an $M$ such that
\be\sup_{\bs{Z}^{(n)}_m} \| \widehat{\bs{\beta}}(\lambda_1, \lambda_2,\gamma, \bs{Z}^{(n)}_m)\|_2 <M<\infty, \label{1-inequality.eqn}\ee
where $\bs{Z}^{(n)}_m$ stands for any contaminated data set by replacing m points in the original data set $\bs{Z}^{(n)}$ with $m$ arbitrary points in $\R^p$.
We seek a contradiction now.
\vs
Replacing $m$ original points $\bs{Z}_i$s with the point
$((\delta, 0,\cdots, 0), \delta\kappa)'$. Denote the contaminated data set by $\bs{Z}^{(n)}_m$ and the estimator based on it as $\widehat{\bs{\beta}}:=\widehat{\bs{\beta}}(\lambda_1, \lambda_2,\gamma, \bs{Z}^{(n)}_m)$. 
 \vs

Let $M_y=\max_{i}|y_i|, M_x=\max_{i}|\bs{x}_{i1}|$.
Let $\bs{\beta}_{\kappa}=(0, \kappa, 0, \cdots, 0)'\in \R^p$ and set $\kappa=M+1$.
Select a large $\delta$ such that   
 $$\delta^2\geq \frac{\max(k-m, k-1)}{n}(M_y+\kappa M_x)^2+\lambda_1\kappa^{\gamma}+\lambda_2 \kappa^2+1,$$
 This is possible since $x^2\to \infty$ when $|x|\to \infty$. Note that $k\geq m$. \vs
It is readily seen that all residuals based on $\bs{\beta}_{\kappa}$ and $m$ contaminated points are zeros. All non-zero residuals correspond to uncontaminated original points.
 Then in the case that $w_i=\mathds{1}(r^2_i\leq r^2_{k:n})$
\begin{align}
Q(\widehat{\bs{\beta}})\leq Q(\bs{\beta}_{\kappa})&=\frac{1}{n}\sum_{i=1}^n r^2_iw_i+\lambda_1\kappa^{\gamma}+\lambda_2 \kappa^2 \nonumber\\[1ex]
 &= \left\{
 \begin{array}{ll}
 \frac{1}{n}\sum_{i=1}^{k-m}r^2_{j_i}+\lambda_1\kappa^{\gamma}+\lambda_2 \kappa^2, & ~~\mbox{if $k>m$}\\[2ex]
 \lambda_1\kappa^{\gamma}+\lambda_2 \kappa^2& ~~\mbox{else},
 \end{array}
 \right.
 \label{1st-inequlity-1.eqn}
\end{align}
where the last equality  
is due that fact that the objective function sums the smallest $k$ squared residuals, but among $n$ squared residuals, $m$ of them are zeros.
\vs For the case $w_i=\mathds{1}(D(r_i, R^{(n)})\leq \alpha)$, one has
\begin{align}
Q(\widehat{\bs{\beta}})\leq Q(\bs{\beta}_{\kappa})&=\frac{1}{n}\sum_{i=1}^n r^2_iw_i+\lambda_1\kappa^{\gamma}+\lambda_2 \kappa^2 \nonumber\\[1ex]
 &= \frac{1}{n}\sum_{i=1}^{k-1}r^2_{j_i}+\lambda_1\kappa^{\gamma}+\lambda_2 \kappa^2,
 \label{1st-inequlity-2.eqn}
\end{align}
 where the last equality is due the fact that there is at most $n-m=k-1$ non-zero residuals.
 Overall we have
 \be
 Q(\widehat{\bs{\beta}})\leq Q(\bs{\beta}_{\kappa})\leq \frac{\max(k-m, k-1)}{n}(M_y+\kappa M_x)^2+\lambda_1\kappa^{\gamma}+\lambda_2 \kappa^2\leq \delta^2-1. \label{2nd-inequality.eqn}
 \ee
\vs
\noin
 On the other hand, for any $\bs{\beta}$ with $\beta_1\leq \|\bs{\beta}\|_2\leq \kappa-1$,
 one has
 \begin{align}
 O(\bs{\beta})&\geq (\kappa\delta-\delta\beta_1)^2 \geq \delta^2,\label{3nd-inequality-1.eqn}
\end{align}
where we utilize the fact that among the $k$ squared residuals, there is at least  one residual that is based on a contaminated point since un-contaminated points are at most $k-1$.
Combining (\ref{2nd-inequality.eqn}) and (\ref{3nd-inequality-1.eqn}) leads to
$$\|\widehat{\bs{\beta}}(\lambda_1, \lambda_2,\gamma, \bs{Z}^{(n)}_m)\|_2 >\kappa-1= 
M,   $$
which contradicts (\ref{1-inequality.eqn}). \hfill \pend
\vs
\noindent
\tb{Proof of Lemma 5.1}
\vs
\noindent
\tb{Proof}: Employing the true model assumption: $\bs{Y}=\bs{X}\bs{\beta}_0+\bs{e}$ and that $\widehat{\bs{\beta}}^n$ is the minimizer  of the RHS of (4.19), 
this is straightforward by some algebraic derivations. \hfill \pend
\vs
\noindent
\tb{Proof of Lemma 5.2}
\vs
\noindent
\tb{Proof}: According to the Definition 1.2 of \cite{RH17}, each $e^*_i $ is a sub-Gaussian variable. Write $\bs{v}^{(j)}=(v_1,\cdots, v_n)':=\bs{x}^{(j)}/c_x$, then $(\bs{e}^*)' \bs{x}^{(j)}/c_x=\sum_{i\in I(\widehat{\bs{\beta}}^n)} v_ie_i$ with $\|\bs{v}\|_2\leq 1$. Following the proof of Theorem 1.6 of \cite{RH17}, one obtains the desired result (i). (ii) follows from the fact that $e_i^2/\sigma^2$ has a $\chi^2$ distribution with one degree of freedom.
\hfill \pend
\vs
\noindent
\tb{Proof of Lemma 5.3}
\vs
\noin
\tb{Proof}: First we note that
\bee
P\Big(\max_{1\leq j\leq p}|(\bs{e}^*)'\bs{x}^{(j)}|/n>q_1/2\Big)&\leq& P\Big(\max_{\|\bs{v}\|\leq 1}|\bs{v}'\bs{e}^*|>nq_1/(2c_x)\Big),
\ene
where $\bs{v}\in \R^n$. Now invoking Lemma 5.2 and Theorem 2.2.2 and Remark 2.2.2 of \cite{P20} (set $nq_1/(2c_x)$ to be the $t$ in Remark 2.2.2), one gets that
$$
P\Big(\max_{1\leq j\leq p}|(\bs{e}^*)'\bs{x}^{(j)}|/n>q_1/2\Big)\leq P\Big(\max_{\|\bs{c}\|\leq 1}|\bs{c}'\bs{e}^*|>q_1/2\Big)\leq \delta/2,
$$
the statement about $P(\mathscr{S}_1)$ follows. \vs
For the statement about $P(\mathscr{S}_2)$, we first invoke Lemma 5.2 and notice that $\|\bs{e}\|^2_{D^*}/\sigma^2$ follows a ${\chi}^2$ distribution with $N_d$ degrees of freedom,  then invoke Lemma 1 and Comments on page 1325 of \cite{LM00}, 
 we get
$$
P\Big(\|\bs{e}\|^2_{D^*}/\sigma^2-N_d\geq 2\sqrt{N_dt}+2t \Big)\leq e^{-t},
$$
now if one sets $\delta/2=e^{-t}$, that is $t=\log 2/\delta$, then one gets that
$$P(\mathscr{S}_2)\geq 1-\delta/2. $$
The second statement follows. \hfill \pend
\vs

\end{document}